\documentclass[journal, final]{IEEEtran}

\usepackage{graphicx}
\usepackage{url}
\graphicspath{{./fig/}}
\usepackage{subfigure}
\usepackage{stackengine}
\usepackage{multirow}
\usepackage{amsmath,amsfonts,amssymb,amsthm}
\usepackage{url}
\usepackage{floatrow}
\usepackage{enumerate}
\usepackage{dsfont}
\usepackage{enumerate}
\usepackage{wrapfig}
\usepackage{epstopdf}

\usepackage{color}
\newfloatcommand{capbtabbox}{table}[][\FBwidth]

\newcommand{\ie}{i.e.}

\newcommand{\MAE}{{\rm MAE}}
\newcommand{\MSE}{{\rm MSE}}
\newcommand{\MRE}{{\rm MRE}}
\usepackage{algorithm}
\usepackage{algorithmic}

\newtheorem{prop}{Proposition}
\newtheorem{thm}{Theorem}

\usepackage{array}
\usepackage{multirow}

\ifCLASSOPTIONcompsoc
\usepackage[nocompress]{cite}
\else
\usepackage{cite}
\fi

\begin{document}
%
\title{Active Regression with Adaptive Huber Loss}
%
%
%

\author{Jacopo Cavazza and Vittorio Murino,~\IEEEmembership{Senior Member,~IEEE}%
\thanks{Jacopo Cavazza and Vittorio Murino are with Pattern Analysis \& Computer Vision, Istituto Italiano di Tecnologia (IIT), Via Morego, 30, 16163, Genova, Italy.}
\thanks{Jacopo Cavazza is also with Dipartimento di Ingegneria Navale, Elettrica, Elettronica e delle Telecomunicazioni (DITEN), University of Genova, Via Opera Pia 11, 16145, Genova, Italy.}
\thanks{Vittorio Murino is also with Computer Science Department, University of Verona, Strada Le Grazie 15, 37134, Verona, Italy.}
\thanks{Primary email contact: \texttt{jacopo.cavazza@iit.it}.}
}

%
%

\markboth{ }
{Jacopo Cavazza and Vittorio Murino: Active Regression with Adaptive Huber Loss}
%



\maketitle


%
\IEEEpeerreviewmaketitle

\begin{abstract}
	This paper addresses the scalar regression problem through a novel solution to exactly optimize the Huber loss in a general semi-supervised setting, which combines multi-view learning and manifold regularization. We propose a principled algorithm to 1) avoid computationally expensive iterative schemes while 2) adapting the Huber loss threshold in a data-driven fashion and 3) actively balancing the use of labelled data to remove noisy or inconsistent annotations at the training stage. In a wide experimental evaluation, dealing with diverse applications, we assess the superiority of our paradigm which is able to combine robustness towards noise with both strong performance and low computational cost.
\end{abstract}

\section{Introduction}

Regression is one of the most widely studied problems in different research disciplines such as statistics, econometrics, computational biology and physics to name a few, and is a pillar topic in machine learning. Formally, it addresses the problem of inferring a functional input-output relationship. Under a classical machine learning perspective, the latter is learnt by means of (training) examples and, to this aim, two mainstream approaches pop up: optimization-based and Bayesian frameworks. In the former, once a suitable hypothesis space is fixed, the goal is minimizing an objective functional, where a loss measures how good the learned regressor reproduces the relationship between input and output variables inside the training set. Instead, in the Bayesian formalism, a prior distribution constrains the solution upon some a priori knowledge, while the final regression map maximizes the posterior/likelihood probability distribution. 

Actually, a problem affecting both paradigms lies in the scarcity of annotations in the training data, which can seriously impact on the generalizability of the regression method. Nevertheless, labelled data are always time-consuming, frequently onerous and sometimes difficult to obtain. Thus, in this respect, semi-supervised approaches play a substantial role in exploiting unlabelled samples to support the search for the solution. Among the most effective semi-supervised algorithms, multi-view learning \cite{S&R:ICML08,Rosenberg:09} considers the structure of the input data as composed by several ``views'' which are associated to different hypothesis spaces employed to construct as many decision functions, finally fused together. Also, manifold regularization is another state-of-the-art semi-supervised method where the geometry of the data space shapes the solution, typically by means of the graph Laplacian operator \cite{Belkin:2006,Minh:2011}.

However, even in the small labelling regime, since annotations are usually provided by human operators, they are frequently prone to errors and noisy in general, making them rather misleading. Hence, for both theoretical and practical aspects, it is of utmost importance to devise algorithms which are able to automatically analyse the data as to guarantee robustness towards outliers. In the literature, several works have tackled such a problem \cite{DeLaTorre}: since the archetypal work \cite{Huber:1964}, many robust regression and classification frameworks \cite{Mangasarian:00,Ando:05,LL:11,Khan:13} successfully leveraged on the (convex and differentiable) Huber loss function

\begin{equation}\label{eq:HuberLoss}
H_\xi \colon \mathbb{R} \to [0,+\infty), \quad H_\xi(y) = \begin{cases} \frac{y^2}{2} & \mbox{if} \; |y| \leq \xi \\ \xi|y| - \frac{\xi^2}{2} & \mbox{otherwise,}
\end{cases}
\end{equation}

where $\xi > 0$. However, as the major drawback of $H_\xi$, there is no closed-form solution to optimize it and, as a consequence, iterative schemes (such as quadratic programming \cite{Ando:05} or self-dual minimization \cite{LL:11}) were previously exploited for either the original Huber loss \cite{Mangasarian:00,LL:11} or its spurious versions (hinge-Huber \cite{Ando:05} or the huberized Laplacian \cite{Khan:13}). Moreover, in all cases, additional computational efforts have to be spent in order to fix the threshold $\xi$, such as statistical efficiency analysis \cite{Mangasarian:00}.

In this work we face all the aforementioned issues trough the following main contributions.
\begin{enumerate}[$I$.]
	\item We derive a novel theoretical solution to exactly optimize the Huber loss in a general multi-view, semi-supervised and manifold regularized setting \cite{Minh:2013}, in order to guarantee a broad applicability of the developed formalism.
	\item We devise the novel Huber Loss Regression (HLR) algorithm to efficiently implement the proposed solution and avoid classical iterative schemes \cite{Ando:05,LL:11}. Moreover, two additional aspects are notable. \\
	{\bf Active-learning}. While taking advantage of the both labelled and unlabelled training samples, the former ones are inspected so that HLR automatically removes those annotations which violate a specific numerical check, whenever recognized as either noisy or inconsistent for learning the regressor. \\
	{\bf Adaptive threshold}. Differently from all \cite{Mangasarian:00, Ando:05, LL:11,  Khan:13}, HLR automatically learns $\xi$ in a data-driven fashion without increasing the computational complexity of the whole pipeline.
	\item Throughout an extensive empirical evaluation, we validate the proposed technique, which allows to score competitive results in curve fitting, learning with noisy labels, classical regression problems and crowd counting application. While using variegate types of data and addressing diverse problems, HLR is able to outperform state-of-the-art regression algorithms.
\end{enumerate}

The paper is outlined as follows. In Sections \ref{sez:sssmvr} and \ref{sez:sol}, we present our exact optimization for the Huber loss after defining the general semi-supervised setting we consider. Section \ref{sez:alg} broadly discusses the HLR algorithm. As to prove its versatility, once benchmarked with a state-of-the-art convex solver in Section \ref{sez:synt}, we registered a strong performance when applying HLR to curve fitting and learning with noisy labels (Section \ref{ssez:active}), classical regression problems (Section \ref{sez:altri}) and crowd counting application (Section \ref{sez:CC}). Finally, Section \ref{sez:end} draws the conclusions.

\section{Multi-view scalar regression}\label{sez:sssmvr}

In this Section, we introduce the formalism to model data points sampled from a composite input space $\mathcal{X}$ which is divided into multiple substructures. Precisely, we assume that for any $\mathbf{x} \in \mathcal{X},$ we have $\mathbf{x} = [x^1,\dots,x^m]$ and $x^\alpha$ belongs to the subspace $\mathcal{X}^\alpha,$ for any $\alpha=1,\dots,m$. This is a very natural way to model high dimensional data: $\mathbf{x}$ is the concatenation of $x^1,\dots,x^m,$ each one representing a particular class of features, that is one out of multiple views \cite{S&R:ICML08,Rosenberg:09,Minh:2013} in which data may be structured. 

In order to find the regression map, we assume that it belongs to an hypothesis space $\mathcal{H}$ of functions $h \colon \mathcal{X} \rightarrow \mathbb{R}$ whose construction is investigated below. For any $\alpha=1,\dots,m,$ let $\kappa^\alpha \colon \mathcal{X}^\alpha \times \mathcal{X}^\alpha \rightarrow \mathbb{R}$ a Mercer kernel \cite{Sc:2002}, that is a symmetric and positive semi-definite function. Let us define $K(\mathbf{x},\mathbf{z}) = {\rm diag}(\kappa^1(x^1,z^1),\dots,\kappa^m(x^m,z^m)) \in \mathbb{R}^{m \times m},$ where $\mathbf{x}=[x^1,\dots,x^m],$ $\mathbf{z}=[z^1,\dots,z^m] \in \mathcal{X}.$ Consider $\mathcal{S}_0$ the space of functions $\mathbf{z} \mapsto f(\mathbf{z}) = \sum_{i = 1}^n K(\mathbf{z},\mathbf{x}_i)u_i$ with $\mathbf{x}_1,\dots,\mathbf{x}_n \in \mathcal{X}$ and $u_1,\dots,u_n$ column vectors in $\mathbb{R}^m.$ Define the norm $\|f\|_K^2 = \sum_{i,j=1}^n u_i^\top\hspace{-.1 cm}K(\mathbf{x}_i,\mathbf{x}_j)u_j.$ The reproducing kernel Hilbert space (RKHS) $\mathcal{S}_K$ related  to $K$ is the completion of $\mathcal{S}_0$ with the limits of Cauchy sequences converging with respect to $\|\cdot\|_K$ \cite{Carmeli:2006}. Finally, $c \in \mathbb{R}^m$ induces a sampling operator $c^\top\hspace{-.1 cm}\colon \mathcal{S}_K \to \mathcal{H}$ whose image is our final hypothesis space. Consequently,
\begin{equation}\label{eq:h}
	h(\mathbf{z}) = c^\top\hspace{-.1 cm}f(\mathbf{z}) = \sum_{i=1}^n c^\top\hspace{-.1 cm}K(\mathbf{z},\mathbf{x}_i) u_i
\end{equation}
is the general expression for every $h \in \mathcal{H}$ with $f \in \mathcal{S}_0.$

{\em Remark.} Differently from other previous multi-view learning paradigms \cite{Rosenberg:09,S&R:ICML08,Minh:2013}, the kernel matrix $K$ is fixed to be diagonal. Empirically, it allows to dedicate an independent kernel to each view, gathering the subspace information codified by our data. In addition, such choice provides some theoretical advantages. Indeed, as pointed out in \cite{Minh:2013}, it is not trivial to prove that the kernel adopted by \cite{Rosenberg:09} is positive semi-definite, while such property is trivially fulfilled in our setting. Also, thanks to \eqref{eq:h}, we can recover $h(\mathbf{z})=[h^1(z^1),\dots,h^m(z^m)]$ for every $m$ by means of the analytical expression
$h^\alpha(z^\alpha) = \sum_i c^\alpha \kappa^\alpha(z^\alpha,x_i^\alpha)u_i^\alpha$, for $\alpha = 1,\dots,m$. Differently, in \cite{S&R:ICML08}, an analogous result holds for $m=2$ only.

\section{Huber multi-view regression with manifold regularization}\label{sez:sol} 

Once the hypothesis space $\mathcal{H}$ is defined according to Section \ref{sez:sssmvr}, we can perform optimization to learn for the regression map. To this aim, we consider a training set $\mathbf{D}$ made of $\ell$ labelled instances $(\mathbf{x}_1,y_1),\dots,(\mathbf{x}_\ell,y_\ell) \in \mathcal{X} \times \mathbb{R}$ and $u$ additional unlabelled inputs $\mathbf{x}_{\ell+1},\dots,\mathbf{x}_{\ell+u} \in \mathcal{X}.$ Among the class of functions \eqref{eq:h}, similarly to \cite{Belkin:2006, Minh:2013}, the regression map is found by minimizing the following objective 
\begin{equation}\label{eq:probl}
	J_{\lambda,\gamma}(f) = \dfrac{1}{\ell} \sum_{i=1}^\ell H_\xi(y_i - c^\top\hspace{-.1 cm} f(\mathbf{x}_i))  + \lambda \|f\|_K^2 + \gamma \|f\|^2_{M}.
\end{equation} 
In equation \eqref{eq:probl}, $\frac{1}{\ell} \sum_{i=1}^\ell H_\xi(y_i - c^\top\hspace{-.1 cm}f(\mathbf{x}_i))$ represents the empirical risk and is defined by means of the Huber loss \eqref{eq:HuberLoss}: it measures how well $c^\top\hspace{-.1 cm} f(\mathbf{x}_i)$ predicts the ground truth output $y_i.$ For $\lambda > 0$, to avoid either under-fitting or over-fitting, $\|f\|_K^2$ is a Tichonov regularizer which controls the complexity of the solution. Instead, $\gamma > 0$ weighs
\begin{equation}
	\|f\|^2_{M} = \sum_{\alpha = 1}^{m}\sum_{i,j=1}^{u+\ell} f^\alpha(x^\alpha_i) M^\alpha_{ij} f^\alpha(x^\alpha_j), 
\end{equation}
which infers the geometrical information of the feature space. The family $\{M^1,\dots,M^m\}$ of symmetric and positive semi-definite matrices $M^\alpha \in \mathbb{R}^{(u+\ell) \times (u+\ell)}$ specifies $\|f\|^2_{M}$ and ensures its non-negativeness. This setting generalizes \cite{Belkin:2006,S&R:ICML08,Rosenberg:09} where $M^1 = \dots = M^m = L,$ being $L$ is the graph Laplacian related to the $(u+\ell) \times (u+\ell)$ adjacency matrix $W.$ The latter captures the interdependencies between $f(\mathbf{x}_1),\dots,f(\mathbf{x}_{u+\ell})$ by measuring their mutual similarity. Then, in this simplified setting, the regularizer rewrites $\sum_{i,j=1}^{u+\ell} w_{ij} \|f(\mathbf{x}_i) - f(\mathbf{x}_j)\|_2^2$ if $w_{ij}$ denotes the $(i,j)$-th entry of $W.$ In this particular case, we retrieve the idea of manifold regularization \cite{Belkin:2006} to enforce nearby patterns in the high-dimensional RKHS to share similar labels, so imposing further regularity on the solution. Conversely, with our generalization, we can consider $m$ different graph Laplacians $L^1,\dots,L^m$ and adjacency matrices $W^1,\dots,W^m$, each of them referring to one view separately. Thus, 
\begin{equation}
	\|f\|^2_{M} = \sum_{\alpha=1}^{m}\sum_{i,j=1}^{u+\ell} w^\alpha_{ij} \left|f^\alpha(x^\alpha_i) - f(x^\alpha_j)\right|^2
\end{equation}
forces the smoothness of $f$ on any subspace $\mathcal{X}^\alpha$ instead of doing it globally on $\mathcal{X}$ as in \cite{Belkin:2006}.

From \eqref{eq:probl}, usual single-viewed ($m=1$), fully supervised ($u = 0$) and classical regularization frameworks ($\gamma = 0$) can be retrieved as particular cases, making our theoretical contribution applicable to a broad field of optimization problems. Precisely, in such general framework, we will now provide a novel solution to guarantee the exact optimization for the Huber loss $H_\xi$ \cite{Huber:1964,Boyd:2004}. Actually, the latter generalizes both the quadratic loss and the absolute value, which can be recovered in the extremal cases $\xi \to +\infty$ and $\xi \to 0$ respectively. $H_\xi$ has been shown to be robust against outliers \cite{Huber:1964} since, in a neighbourhood of the origin, it penalizes small errors in a more smoother way than absolute value, whereas, when $|y| \geq \xi,$ the linear growth plays an intermediate role between over-penalization (quadratic loss) and under-penalization (absolute value) of bigger errors. Globally, it resumes the positive aspect of the two losses, while remarkably mitigating their weaknesses. Nevertheless, $H_\xi$ requires the threshold $\xi$ to be set: to this end, we devise an adaptive pipeline to learn it from data (see Section \ref{sez:alg}) as to mitigate such drawback and exploit the robustness of the Huber loss.

In order to optimize our objective \eqref{eq:probl}, we can exploit the theory of the RKHS and the Representer Theorem \cite{Minh:2011}. Precisely, in addition to ensuring existence and uniqueness of the minimizer $f^\star = \arg \min_{f \in \mathcal{S}_k} J_{\lambda,\gamma}(f)$, the latter provides the following expansion valid for any $\mathbf{z} \in \mathcal{X}$:
\begin{equation}\label{eq:exp}
	f^\star(\mathbf{z}) = \sum_{j=1}^{u+\ell} K(\mathbf{z},\mathbf{x}_j)w_j,
\end{equation}
which involves the data inputs $\mathbf{x}_1,\dots,\mathbf{x}_{u+\ell}$ and some $w_1,\dots,w_{u+\ell} \in \mathbb{R}^m.$ Equation \eqref{eq:exp} is the main tool exploited to deduce our general solution through the following result.

\begin{thm}[General solution for Huber loss multi-view manifold regularization regression]\label{th:th}
	For any $\xi > 0,$ the coefficients $\mathbf{w} = [w_1,\dots,w_{u+\ell}]^\top$ defining the solution \eqref{eq:exp} of problem \eqref{eq:probl} are given by
	\begin{align}
		& 2\ell \lambda w_i + 2 \ell \gamma \sum_{j,h=1}^{u+\ell} \textsc{M}_{ij} K(\mathbf{x}_j,\mathbf{x}_h)w_h = \label{eq:HSMVL} \nonumber \\ & =\begin{cases}
			-\xi c & \mbox{if} \; i \in L_+[\mathbf{D},\mathbf{w},\xi] \\
			\left(y_i - \displaystyle \sum_{j=1}^{u+\ell} c^\top\hspace{-.1 cm}K(\mathbf{x}_i,\mathbf{x}_j) w_j \right) c & \mbox{if} \; i \in L_0[\mathbf{D},\mathbf{w},\xi] \\
			+\xi c & \mbox{if} \; i \in L_-[\mathbf{D},\mathbf{w},\xi] \\
			0 & \mbox{otherwise,}   
		\end{cases}
	\end{align}
	where $\lambda,\gamma > 0$ and we set $\textsc{M}_{ij} = {\rm diag}(M^1_{ij},\dots,M^m_{ij}),$
	\begin{align*}  
		L_+[\mathbf{D},\mathbf{w},\xi] &= \left\{ i \leq \ell \colon \sum_{j=1}^{u+\ell} c^\top\hspace{-.1 cm} K(\mathbf{x}_i,\mathbf{x}_j)w_j \geq y_i + \xi \right\}\hspace{-.1cm},\\ 
		L_0[\mathbf{D},\mathbf{w},\xi] &= \left\{ i \leq \ell \colon \left|\sum_{j=1}^{u+\ell} c^\top\hspace{-.1 cm}K(\mathbf{x}_i,\mathbf{x}_j)w_j - y_i \right| < \xi \right\}\hspace{-.1cm},\\		 
		L_-[\mathbf{D},\mathbf{w},\xi] &= \left\{ i \leq \ell \colon \sum_{j=1}^{u+\ell} c^\top\hspace{-.1 cm}K(\mathbf{x}_i,\mathbf{x}_j)w_j \leq y_i - \xi \right\}\hspace{-.1cm}.\\ 
	\end{align*} 
\end{thm}

\proof To not fragment the dissertation, the proof was moved to the \textbf{\textsc{Appendix}}.
\endproof

Let us interpret numerically the sets $L_-[\mathbf{D},\mathbf{w},\xi],$ $L_0[\mathbf{D},\mathbf{w},\xi],$ and $L_+[\mathbf{D},\mathbf{w},\xi].$ First of all, they provide a partition of $\{1,\dots,\ell\}.$ Define $\varepsilon_i = \left|\sum_{j=1}^{u+\ell} c^\top\hspace{-.1 cm}K(\mathbf{x}_i,\mathbf{x}_j)w_j - y_i\right|,$ the absolute in-sample error involving the $i$-th labelled element $(\mathbf{x}_i,y_i) \in \mathbf{D}.$ Call $\varepsilon_\infty =  \max_{i \leq \ell} \varepsilon_i$ their maximum. Thus, the set $L_0[\mathbf{D},\mathbf{w},\xi] = \{ i \leq \ell \colon \varepsilon_i < \xi\}$ collects all the indexes of those instances which are more effective in strictly reducing absolute errors $\varepsilon_i$ below threshold $\xi.$ Instead, in $L_+[\mathbf{D},\mathbf{w},\xi]$ and $L_-[\mathbf{D},\mathbf{w},\xi],$ the target variables are over and under-estimated, respectively. Furthermore, when $L_0[\mathbf{D},\mathbf{w},\xi] = \{1,\dots,\ell\}$, since $\varepsilon_i < \xi$ for any $i,$ we obtain $\varepsilon_\infty < \xi.$ So, if $L_+[\mathbf{D},\mathbf{w},\xi]$ and $L_-[\mathbf{D},\mathbf{w},\xi]$ are empty, $\xi$ is an upper bound for the absolute error of the labelled training set in $\mathbf{D}$. In spite of the nice numerical interpretation, the sets $L_+[\mathbf{D},\mathbf{w},\xi], L_0[\mathbf{D},\mathbf{w},\xi], L_-[\mathbf{D},\mathbf{w},\xi]$ can be computed only if $\mathbf{w}$ is given. Thus, apparently, we can not exploit the general solution \eqref{eq:HSMVL} directly implementing it. Actually, such issue is solved in Section \ref{sez:alg}.

\section{The HLR algorithm}\label{sez:alg}

As a preliminary step to introduce the algorithm to minimize the objective \eqref{eq:probl}, we rephrase \eqref{eq:HSMVL} into the matrix formulation $\left(\mathbf{C}^\star \mathbf{C} \mathbf{Q}[\mathbf{D},\mathbf{w},\xi]+ 2\ell \lambda \mathbf{M} \mathbf{K}[\mathbf{x}]\right) \mathbf{w} + 2\ell \gamma \mathbf{w} = \mathbf{C}^\star \mathbf{b}[\mathbf{D},\mathbf{w},\xi],$ exploiting the notation reported below. 
\begin{itemize}
	\item $\mathbf{w} = [w_1,\dots,w_{u+\ell}]^\top$ collects $ w_j \in \mathbb{R}^m$ for any $j.$
	\item Let $\mathbf{y} = [y_1,\dots,y_\ell]^\top$ and denote with $\mathbf{y}_0$ the result of appending to $\mathbf{y}$ a $u \times 1$ vector of zeros.
	\item $\mathbf{M}$ is the block matrix collecting $\textsc{M}_{ij}$ for any $i,j.$
	\item Consider $\mathbf{K}[\mathbf{x}],$ the Gram matrix defined by $K_{ij}[\mathbf{x}]=K(\mathbf{x}_i,\mathbf{x}_j)$ for $i,j=1,\dots,u+\ell.$
	\item Denoting $I_n$ the $n \times n$ identity matrix and $\otimes$ the Kronecker tensor product, let $\mathbf{C} = I_{u+\ell} \otimes c^\top$ and $\mathbf{C}^\star = I_{u+\ell} \otimes c.$
	\item Let $\mathbf{Q}[\mathbf{D},\mathbf{w},\xi] \in \mathbb{R}^{m(u+\ell) \times m(u+\ell)}$ a block matrix,  $$Q_{ij}[\mathbf{D},\mathbf{w},\xi] = \begin{cases} K_{ij}[\mathbf{x}] & \mbox{if} \; i \in L_0[\mathbf{D},\mathbf{w},\xi] \\ 0 & \mbox{otherwise.} \end{cases}$$
	\item $\mathbf{b}[\mathbf{D},\mathbf{w},\xi] = [b_1[\mathbf{D},\mathbf{w},\xi], \dots, b_{u+\ell}[\mathbf{D},\mathbf{w},\xi]]^\top,$ 
	$$b_i[\mathbf{D},\mathbf{w},\xi] = \begin{cases}
	-\xi & \mbox{if} \; i \in L_+[\mathbf{D},\mathbf{w},\xi] \\ y_i & \mbox{if} \; i \in L_0[\mathbf{D},\mathbf{w},\xi] \\ +\xi & \mbox{if} \; i \in L_-[\mathbf{D},\mathbf{w},\xi] \\ 0 & \mbox{if} \; i = \ell+1,\dots,u+\ell.
	\end{cases}$$
\end{itemize}

\begin{algorithm}[t!]
	\caption{HLR algorithm pseudo-code}
	\label{alg:LiM}
	\begin{algorithmic}
		\STATE {\bfseries Input:} $\mathbf{D}$ dataset, $M^1,\dots,M^m$ positive definite $m \times m$ matrices, $\lambda > 0$ Tichonov regularizing parameter, $\gamma > 0$ manifold regularization parameter, $\Delta \xi > 0$ updating rate, maximum number $T$ of refinements.
		\STATE {\bfseries Output:} Coefficients vector $\mathbf{w}^\star = [w_1^\star,\dots,w_{u+\ell}^\star]^\top.$
		
		\STATE Find $\mathbf{w}^{(0)}$ solving $\left(\mathbf{C}^\star \mathbf{C} + \ell \gamma \mathbf{M} \right)  \mathbf{K}[\mathbf{x}]\mathbf{w}^{(0)} + \ell \lambda \mathbf{w}^{(0)} = \mathbf{C}^\star \mathbf{y}_0.$ 
		\STATE $\mathbf{D}^{(0)} := \mathbf{D}$ and compute $\xi^{(0)}$ as in \eqref{eq:xitau}.
		\FOR{$\tau = 1,\dots,T$}
		\STATE $\widetilde{\xi}^{(\tau)} := \xi^{(\tau - 1)} - \Delta \xi.$ 
		\STATE Solve \eqref{eq:in} in the unknown $\mathbf{w}^{\rm new} := \widetilde{\mathbf{w}}^{(\tau)},$ with $\mathbf{w}^{\rm old} := \mathbf{w}^{(\tau-1)},$  $\mathbf{D}:=\mathbf{D}^{(\tau-1)}$ and $\xi := \widetilde{\xi}^{(\tau)}.$
		\STATE Compute the sets $L_0,L_+$ and $L_-$ on $[\mathbf{D}^{(\tau - 1)},\widetilde{\mathbf{w}}^{(\tau)},\widetilde{\xi}^{(\tau)}].$
		\IF{$L_0[\mathbf{D}^{(\tau - 1)},\widetilde{\mathbf{w}}^{(\tau)},\widetilde{\xi}^{(\tau)}]$ is empty}
		\STATE Return $\mathbf{w}^\star := \mathbf{w}^{(\tau-1)}.$
		\ELSE
		\STATE Obtain $\mathbf{D}^{(\tau)}$ from $\mathbf{D}^{(\tau)}$ by removing the labels $y_i$ of the data points $(\mathbf{x}_i,y_i) \in \mathbf{D}^{(\tau-1)}$ such that $i \in L_+[\mathbf{D}^{(\tau - 1)},\widetilde{\mathbf{w}}^{(\tau)},\widetilde{\xi}^{(\tau)}] \cup L_-[\mathbf{D}^{(\tau - 1)},\widetilde{\mathbf{w}}^{(\tau)},\widetilde{\xi}^{(\tau)}].$			
		\STATE Sort $\mathbf{w}^{(\tau)}$ permuting $\mathbf{w}^{(\tau-1)}$ in a way that the elements $w_j^{(\tau-1)}$ with $j \in L_0[\mathbf{D}^{(\tau - 1)},\widetilde{\mathbf{w}}^{(\tau)},\widetilde{\xi}^{(\tau)}]$ occupy the first entries.
		\STATE Compute $\xi^{(\tau)}$ using relation \eqref{eq:xitau}.
		\ENDIF
		\ENDFOR
		\STATE Return $\mathbf{w}^\star := \mathbf{w}^{(\tau)}$
	\end{algorithmic}
\end{algorithm}

As observed in Section \ref{sez:sol}, the bottleneck relates to the sets $L_+[\mathbf{D},\mathbf{w},\xi],$ $L_0[\mathbf{D},\mathbf{w},\xi],$ and $L_-[\mathbf{D},\mathbf{w},\xi],$ which are not computable if $\mathbf{w}$ is unknown, compromising the numerical implementability of \eqref{eq:HSMVL}. As a very naive approach, if a certain initialization for the solution is provided, we can therefore set a scheme in which, at each iteration, $L_0,L_+$ and $L_-$ are firstly computed for an initial approximation $\mathbf{w} = \mathbf{w}^{\rm old}$ of the solution. Then, $\mathbf{w}$ is updated to  $\mathbf{w}^{\rm new}$ solving the linear system 
\begin{align}  &\left(\mathbf{C}^\star \mathbf{C} \mathbf{Q}[\mathbf{D},\mathbf{w}^{\rm old},\xi]+ 2\ell \lambda \mathbf{M} \mathbf{K}[\mathbf{x}]\right) \mathbf{w}^{\rm new} +\nonumber \\ & + 2\ell \gamma \mathbf{w}^{\rm new} = \mathbf{C}^\star \mathbf{b}[\mathbf{D},\mathbf{w}^{\rm old},\xi] \label{eq:in}
\end{align} for a fixed $\xi > 0.$ Such naive approach suffers from several troubles. Indeed, an initialization is required for the scheme, whose convergence is eventually not guaranteed. Also, the Huber loss threshold $\xi$ has to be manually selected. 

As a more effective approach, consisting in the main contribution of the paper, we introduce the Huber loss regression algorithm (HLR) to perform a principled optimization and compute the exact solution of the problem \eqref{eq:HSMVL}, while also learning the best value for $\xi$ via sequential refinements $\xi^{(0)},\xi^{(1)},\dots,\xi^{(T)}$ and automatic labelled data scanning. Algorithm \ref{alg:LiM} describes the pseudo-code for HLR, which is discussed in Sections \ref{ssez:HLR} and \ref{sez:AL}, while the following results provide its theoretical foundation. 
\begin{prop}\label{prop:conv}
	For any $\tau = 0,1,\dots,T,$ the coefficients $\mathbf{w}^{(\tau)}$ satisfy \eqref{eq:HSMVL} with $\xi = \xi^{(\tau)},$ where 
	\begin{equation} \label{eq:xitau}
		\xi^{(\tau)} = \max_{i \leq \ell}\left| \sum_{j=1}^{u+\ell} c^\top\hspace{-.1 cm} K(\mathbf{x}_i,\mathbf{x}_j) w_j^{(\tau)} - y_i \right|.
	\end{equation}
\end{prop}

\proof It is enough to show that, for any $\tau=0,1,\dots,T,$ we get $L_0[\mathbf{D}^{(\tau)},\mathbf{w}^{(\tau)},\xi^{(\tau)}] = \{1,\dots,\ell\}.$ Let's go by induction. 

For $\tau = 0,$ as mentioned before, $\mathbf{w}^{(0)}$ solves \eqref{eq:HSMVL} for $\xi=+\infty$ since, for any dataset $\mathbf{D}$ and any vector $\mathbf{w}$ of coefficients $w_1,\dots,w_{u+\ell},$ we have $L_0[\mathbf{D},\mathbf{w},+\infty] = \{1,\dots,\ell\}$. Since $\xi^{(0)}$ represents the maximum absolute error inside the training set $\mathbf{D}^{(0)} = \mathbf{D}$ when the solution of the optimization problem is specified by $\mathbf{w}^{(0)},$ we have $L_0[\mathbf{z},\mathbf{w}^{(0)},\xi^{(0)}] = \{1,\dots,\ell\}$. So the thesis is proved for $\tau = 0.$ 

Now, let's assume that \eqref{eq:xitau} holds for the $(\tau - 1)$-th refinement and we prove it for the $\tau$-th one. As a consequence, 
\begin{equation}\label{eq:ass}
L_0[\mathbf{D}^{(\tau-1)},\mathbf{w}^{(\tau-1)},\xi^{(\tau-1)}] = \{1,\dots,\ell\}
\end{equation}
and we must show that the same relation is valid also for $\tau.$ Once computed $\widetilde{\mathbf{w}}^{(\tau)},$ we do not discard $y_i$ from the training data $\mathbf{D}^{(\tau - 1)}$ if and only if $\left| \sum_{j=1}^{u+\ell} c^\top K(\mathbf{x}_i,\mathbf{x}_j) \widetilde{w}_j^{(\tau)} - y_i \right| \leq \widetilde{\xi}^{(\tau)}.$ Since the algorithm requires to compute $\mathbf{w}^{(\tau)}$ by permuting $\mathbf{w}^{(\tau-1)}$ in a way that the elements $w_j^{(\tau-1)}$ with $j \in L_0[\mathbf{D}^{(\tau - 1)},\widetilde{\mathbf{w}}^{(\tau)},\widetilde{\xi}^{(\tau)}]$ occupy the first entries, we have $\left| \sum_{j=1}^{u+\ell} c^\top K(\mathbf{x}_i,\mathbf{x}_j) w_j^{(\tau)} - y_i \right| \leq \widetilde{\xi}^{(\tau)}$ thanks to the assumption \eqref{eq:ass}. Since $\xi^{(\tau)}$ is defined as the maximum of a finite set of elements all bounded by $\widetilde{\xi}^{(\tau)},$ we conclude 
\begin{equation}\label{eq:min}
\xi^{(\tau)} = \max_{i = 1,\dots,\ell}\left| \sum_{j=1}^{u+\ell} c^\top K(\mathbf{x}_i,\mathbf{x}_j) w_j^{(\tau)} - y_i \right| \leq \widetilde{\xi}^{(\tau)}.
\end{equation}
From the previous relation and from the definition of the set $L_0,$ we obtain $L_0[\mathbf{D}^{(\tau)},\mathbf{w}^{(\tau)},\xi^{(\tau)}] = \{1,\dots,\ell\}$.
\endproof

\begin{prop}\label{prop:decr}
	The sequence $\xi^{(0)},\xi^{(1)},\dots,\xi^{(T)}$ is monotonically strictly decreasing.
\end{prop}

\proof In formul\ae, we want to show that $\xi^{(0)}>\xi^{(1)}>\dots>\xi^{(T)}$. In order to prove monotonicity, we fix an arbitrary refinement $\tau = 1,\dots,T$ and our goal is to show $\xi^{(\tau)} < \xi^{(\tau - 1)}.$ Directly using \eqref{eq:min}, we have
\begin{equation}\label{eq:e1}
\xi^{(\tau)} \leq \widetilde{\xi}^{(\tau)}.
\end{equation} 
By definition of $\widetilde{\xi}^{(\tau)},$ 
\begin{equation}\label{eq:e2}
\widetilde{\xi}^{(\tau)} = \xi^{(\tau - 1)} - \Delta \xi,
\end{equation}
and, since $\Delta\xi >0,$ then 
\begin{equation}\label{eq:e3}
\xi^{(\tau-1)} - \Delta \xi < \xi^{(\tau - 1)}.
\end{equation}
Combining the equations \eqref{eq:e1}, \eqref{eq:e2} and \eqref{eq:e3} we get
\begin{equation}\label{eq:salamella}
\xi^{(\tau)} < \xi^{(\tau - 1)}.
\end{equation}
The thesis follows after the generality of $\tau$ in \eqref{eq:salamella}. 
\endproof

\subsection{Insights about the HLR pseudo-code}\label{ssez:HLR}

Essentially, Algorithm \ref{alg:LiM} solves the burden related to the computability of the sets $L_0,L_+$ and $L_-,$ by adapting the threshold $\xi$ of the Huber loss. Indeed, $\mathbf{w}^0$ is easily computable in the setting $\xi = +\infty$ since $L_0[\mathbf{D},\mathbf{w},+\infty] = \{1,\dots,\ell\}$ for any $\mathbf{w}$ and $\mathbf{D}$ and \eqref{eq:HSMVL} reduces to a ordinary linear system. Precisely, thanks to the definition of $\xi^{(\tau)}$ in \eqref{eq:xitau}, we can ensure that $\mathbf{w}^{(0)}$ solves the problem \eqref{eq:HSMVL} for the value $\xi = \xi^{(0)}$ of the Huber loss threshold. 
The refinements for $\xi$ start with a fixed reduction of $\xi^{(\tau-1)}$ by $\Delta \xi > 0$ and, once \eqref{eq:in} is solved, the final value $\xi^{(\tau)}$ is updated: this is the key passage to ensure that Proposition \ref{prop:conv} holds for any $\tau$. Let us stress again that, \emph{for any refinement $\tau=0,\dots,T,$ the vector $\mathbf{w}^{(\tau)}$ gives the exact solution \eqref{eq:HSMVL} for our robust scalar regression framework \eqref{eq:probl}, where the Huber loss threshold $\xi$ equals to $\xi^{(\tau)}.$} 

Additionally, HLR is able to automatically select which output variables $y_1,\dots,y_\ell$ do not provide sufficient information to learn the regressor. Indeed, at each refinement, HLR scans the labelled training set $\{(\mathbf{x}_1,y_1),\dots,(\mathbf{x}_\ell,y_\ell)\} \subset \mathbf{D}^{(\tau-1)}$, checking whether, for $i = 1,\dots, \ell$,
\begin{equation}\label{eq:co}
	\left| \sum_{j=1}^{u+\ell} c^\top\hspace{-.1 cm}K(\mathbf{x}_i,\mathbf{x}_j) \widetilde{w_j}^{(\tau)} - y_i \right| \geq \widetilde{\xi}^{(\tau)}.
\end{equation} 
Equation \eqref{eq:co} means that, for $\mathbf{x}_i,$ the prediction of HLR is suboptimal, since differing from the actual value $y_i$ for more than $\widetilde{\xi}^{(\tau)}.$ In such case, the algorithm automatically removes $y_i$ from the dataset, assigning $\mathbf{x}_i$ to be unlabelled in $\mathbf{D}^{(\tau)}$ and actually trying to recover a better prediction by looking for the output value in the RKHS which is closest to $f(\mathbf{x}_i)$ in the sense of the norm $\|\cdot\|_M$. For the sake of clarity, notice that the computational cost does not change when some scalar $y_i$ is removed from the training set. Indeed, any unlabelled input $\mathbf{x}_i \in \mathbf{D}^{(\tau)}$ still counts one equation, as it is when $(\mathbf{x}_i,y_i) \in \mathbf{D}^{(\tau - 1)}.$ We will refer to such automatic balance of labelled data as the {\bf active-learning} component of HLR (see Section \ref{sez:AL}).

Furthermore, thanks to Proposition \ref{prop:decr}, the algorithms perform an automatic {\bf threshold adaptation} for $\xi.$ Indeed, $\xi^{(0)} > \xi^{(1)} > \dots > \xi^{(T)}$ is a decreasing sequence whose latter element represents the data-driven selection performed by HLR for the optimal Huber loss threshold $\xi,$ after $T$ refinements. Precisely, according to \eqref{eq:xitau}, such optimality is measured in terms of $T$ successive decreasing reductions of the absolute error paid inside the labelled part of the training set.

Finally, let us conclude with a couple of details. First, the computational cost of HLR is $O((T+1) m^2 (u+ \ell )^2 ).$ Second, once coefficients $w_1^\star,\dots,w_{u+\ell}^\star$ are computed by Algorithm \ref{alg:LiM}, $h^\star(\mathbf{v}) = \sum_{j=1}^{u+\ell} c^\top K(\mathbf{v},\mathbf{x}_j) w^\star_j$ is the predicted output associated to the multi-view test instance $\mathbf{v} \in \mathcal{X}.$ 

\subsection{Interpreting the active-learning component}\label{sez:AL}

The active-learning component of HLR has a natural interpretation in machine learning. Indeed, consider $\mathbf{x}_i$ to represent a collection of different types (\ie, views) of features which are computed on the raw input data by enhancing separated discriminative characteristics, while possibly reducing the noise impact. Conversely, all the labels $y_i$ are directly obtained from ground truth annotations, which are usually acquired by humans and, consequently, more prone to noise and errors. Also, during the learning phase, semantic ambiguities may result in descriptors which are, at the same time, dissimilar in terms of corresponding labels and similar while compared in the feature space. Clearly, such condition violates the classical assumption in semi-supervision where the conditional probability distribution $p(y|\mathbf{x})$ smoothly changes over the geodesics induced by $p_{\mathcal{X}}(\mathbf{x})$ \cite{Belkin:2006}. Thus, despite classical frameworks leveraging on unlabelled examples can be easily damaged in such situation, differently, HLR shows a unique robustness towards corrupted/misleading annotations, being able to totally and automatically discard them.

\section{Empirical validation of HLR}\label{sez:res}

This Section presents our empirical analysis of HLR. In Section \ref{sez:synt}, we compare our algorithm, with a state-of-the-art optimizer for convex objective functions. In Section \ref{ssez:active}, the HLR active-labelling component is applied on noisy curve fitting and benchmarked against several approaches for learning with noisy labels. Section \ref{sez:altri} compares HLR with popular regression methods on classical machine learning datasets. Finally, in Section \ref{sez:CC}, we consider the crowd counting application, validating our method against the state-of-the-art ones in the literature through several experiments on three benchmark datasets.

\begin{table*}[t!]
	\centering
	\caption{Classification accuracies to compare different algorithms against HLR for the task of learning with noisy labels in a binary problem. We considered different rates $\rho_+$ and $\rho_-$ to flip positive and negative examples, respectively. Best results in bold.}
	\label{tab:fff}
	\begin{tabular}{|r|cc|ccccc|c|}
		Dataset & $\rho_+$ & $\rho_-$ & $\widetilde{ \ell }_{\rm log}$ & C-SVM & PAM & NHERD & RP & HLR \\\hline\hline
		& $0.2$ & $0.2$ & 70.12\% & 67.85\% & 69.34\% & 64.90\% & 69.38\% & {\bf 73.86\%} \\
		\emph{Breast Cancer} & $0.3$ & $0.1$ & 70.07\% & 67.81\% & 67.79\% & 65.68\% & 66.28\% & {\bf 71.90\%}	 \\
		& $0.4$ & $0.4$ & {\bf 67.79\%} & {\bf 67.79\%} & 67.05\% & 56.50\% & 54.19\% & 59.12\% \\\hline
		& $0.2$ & $0.2$ & {\bf 76.04\%} & 66.41\% & 69.53\% & 73.18\% & 75.00\% & 75.39\% \\
		\emph{Diabetes} & $0.3$ & $0.1$ & {\bf 75.52\%} & 66.41\% & 65.89\% & 74.74\% & 67.71\% & 74.35\% \\
		& $0.4$ &$ 0.4$ & 65.89\% & 65.89\% & 65.36\% & {\bf 71.09\%} & 62.76\% & 66.37\% \\\hline
		& $0.2$ & $0.2$ & 87.80\% & 94.31\% & {\bf 96.22\%} & 78.49\% & 84.02\% & 92.43\% \\
		\emph{Thyroid} & $0.3$ & $0.1$ & 80.34\% & {\bf 92.46\%} & 86.85\% & 87.78\% & 83.12\% & 85.35\% \\
		& $0.4$ &$ 0.4$ & 83.10\% & 66.32\% & 70.98\% & {\bf 85.95\%} & 57.96\% & 84.15\% \\ \hline
		& $0.2$ & $0.2$ & 71.80\% & 68.40\% & 63.80\% & 67.80\% & 62.80\% & {\bf 75.21\%}   \\
		\emph{German} & $0.3$ & $0.1$ & 71.40\% & 68.40\% & 67.80\% & 67.80\% & 67.40\% & {\bf 72.86\%} \\
		& $0.4$ &$ 0.4$ & 67.19\% & {\bf 68.40\%} & 67.80\% & 54.80\% & 59.79\% & 62.54\% \\ \hline
		& $0.2$ & $0.2$ & {\bf 82.96\%} & 61.48\% & 69.63\% & {\bf 82.96\%} & 72.84\% & 81.53\% \\
		\emph{Heart} & $0.3$ & $0.1$ & {\bf 84.44\%} & 57.04\% & 62.22\% & 81.48\% & 79.26\% & 77.28\% \\
		& $0.4$ &$ 0.4$ & 57.04\% & 54.81\% & 53.33\% & 52.59\% & 68.15\% & {\bf 70.69\%} \\ \hline
		& $0.2$ & $0.2$ & 82.45\% & 91.95\% & 92.90\% & 77.76\% & 65.29\% & {\bf 92.92\%} \\
		\emph{Image} & $0.3$ & $0.1$ & 82.55\% & 89.26\% & 89.55\% & 79.39\% & 70.66\% & {\bf 91.75\%} \\
		& $0.4$ & $ 0.4$ & 63.47\% & 63.47\% & 73.15\% & 69.61\% & 64.72\% & {\bf 82.38\%}  \\ \hline
	\end{tabular}
\end{table*}

\subsection{Comparison with the state-of-the-art convex solver}\label{sez:synt}

As stressed in the Introduction and proved in Section \ref{sez:alg}, HLR leverages on an exact solution for optimizing the Huber loss, as an opposed perspective to iterative solving. In order to experimentally check the powerfulness of such trait, we compare against CVX \cite{cvx2}, the state-of-the-art optimization tool for convex problems. Precisely, by either exploiting HLR or CVX, we are able to optimize the same objective functional \eqref{eq:probl}, consequently investigating which method is more efficient in terms of both reconstruction error and running time. Also, we are able to inspect HLR in the automatic pipeline of learning $\xi$ against a standard cross-validation procedure which is necessary for CVX. 

To do so, we consider the linear regression problem to predict $y \in \mathbb{R}$ from $\mathbf{x} \in \mathbb{R}^{10}$ where $y = \beta^\top \mathbf{x}$,  $\beta = [1/10,\dots,1/10]^\top.$ We randomly generate $n=50,100,500$ and $1000$ samples $\mathbf{x}$ from a uniform distribution over the unit $10$-dimensional hypercube $[0,1]\times\dots\times[0,1].$ As a further experiment, we introduce some outliers to the model which becomes $y = \beta^\top \mathbf{x} + \epsilon,$ where the additive noise $\epsilon$ is distributed according to a zero-mean Gaussian with $0.1$ variance. For HLR, $T=1$ and $\Delta\xi = 0.1$, $\lambda = 10^{-2}$ and $\gamma = 10^{-3}$ are fixed.  The performance of CVX and HLR are measured via the reconstruction error between the ground truth values and the predictions. Also, we monitor the computational running time of both.

\begin{figure}[H]
	\centering
	\includegraphics[width=.325\columnwidth,keepaspectratio]{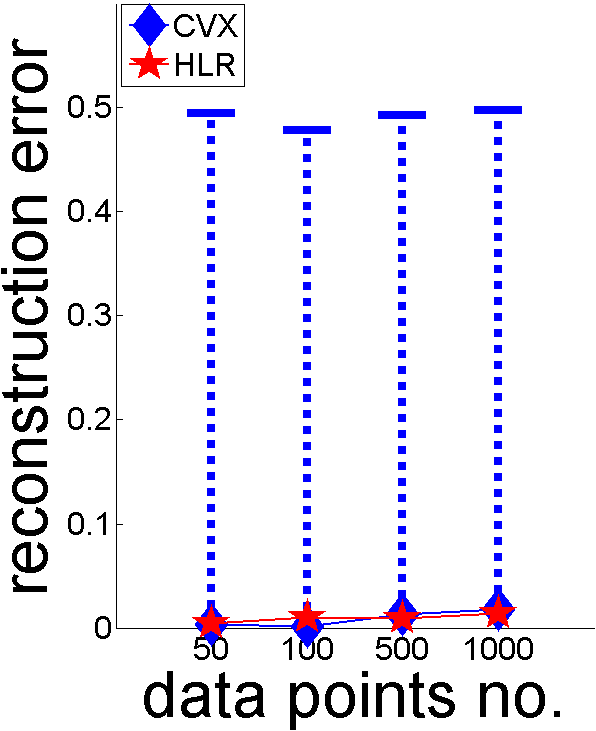} \qquad \qquad
	\includegraphics[width=.325\columnwidth,keepaspectratio]{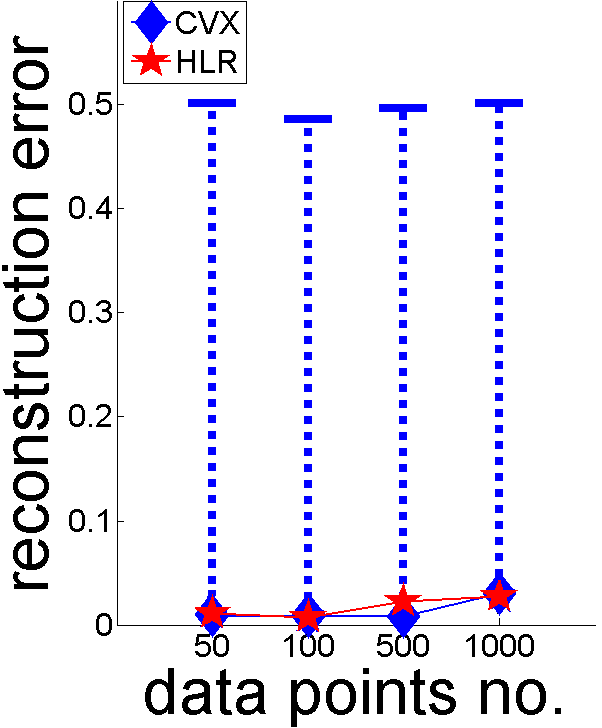}
	\caption{HLR versus CVX \cite{cvx2} in either a noise-free (left) or additive Gaussian noise setup (right). For CVX, after cross validating  $\xi \in \{ 10^{-3},10^{-2},\dots,10^{3} \}$, the blue diamonds represent the best configuration resulting in the lowest reconstruction error, whose fluctuation is represented by means of the error bars.} 
	\label{fig:synt}
\end{figure}

The analysis of Figure \ref{fig:synt} yields to the following comments.

$-$ Numerically, our general solution shows a comparable performance with respect to classical iterative schemes in terms of reconstruction error.\\
$-$ For both algorithms, the noise $\epsilon$ does not remarkably influence the reconstruction error: this is due to the robustness provided by the Huber loss.\\
$-$ Undoubtedly, the reconstruction error of CVX greatly fluctuates when $\xi$ varies in $\{10^{-3},10^{-2},\dots,10^3\}$.

\begin{table}[t!]
	\centering
	\caption{HLR versus CVX. Computational running time (measured in seconds).}
	\label{tab:time}
	\begin{tabular}{|r|c|c|c|c|}
		& $n=50$ & $n=100$ & $n=500$ & $n=1000$ \\ \hline \hline
		CVX & 1.0 \hspace{-.1cm}$\pm$\hspace{-.1cm} 0.2 & 1.9 \hspace{-.1cm}$\pm$\hspace{-.1cm} 0.3 & 36.0 \hspace{-.1cm}$\pm$\hspace{-.1cm} 10.0 & 179.7 \hspace{-.1cm}$\pm$\hspace{-.1cm} 4.2 \\
		HLR & {\bf 0.1} & {\bf 0.1} & {\bf 3.4} & {\bf 18.1} \\\hline
	\end{tabular}
\end{table}

Moreover, Table \ref{tab:time} clarifies that HLR is much faster: the runtime\footnote{For all experiments, we used MATLAB R2015b on a Intel(R) Xeon(R) CPU X$5650$ $@ 2.67$ GHz $\times 2$ cores and $12$ GB RAM.} of HLR is about a few seconds even if $n$ grows while, for CVX, it sharply raises in the cases $n=500$ and $n=1000.$ Also, it is worth nothing that the computational running time for CVX is averaged over all the cross-validating repetitions for the different $\xi$ values, leading to the mean and standard deviation values reported in Table \ref{tab:time}.

Globally, HLR provides low reconstruction errors as CVX, being superior to it in terms of 1) quicker computation and 2) automatic learning $\xi$.

\subsection{Evaluation of the active-learning component}\label{ssez:active}

In this Section, we evaluate the robustness provided by the HLR active-learning component resulted from the usage of the Huber loss. For this purpose, we consider a noisy curve fitting experiment and we also faced the problem of binary classification in a corrupted data regime.

\emph{Noisy curve fitting.} \quad As in Section \ref{sez:synt}, starting from the same linear model $y = \beta^\top\hspace{-.1 cm}\mathbf{x}$, we severely corrupted a random percentage of target data points by inverting their sign. It is a quite sensible change since each entry of $\mathbf{x}$ is uniformly distributed in $[0,1],$ being thus non-negative. Consequently, our algorithm should be able to recognize the negative data as clear outlier and automatically remove them from the training set. Such evaluation is performed through Table \ref{tab:QS} where, for different noise rates, we report the reconstruction error while measuring whether the labels removed by HLR actually refers to corrupted inputs. To do the latter, we employ the S\o{}rensen-Dice index $s$ \cite{SD:45} to measure the amount of corrupted data effectively removed by the HLR. In formul\ae, $s = \frac{2|\mathcal{C} \cap \mathcal{R}|}{|\mathcal{C}|+|\mathcal{R}|}$ where the sets $\mathcal{C}$ and $\mathcal{R}$ collects the corrupted and removed data, respectively: $s \in [0,1]$ and spans from the worst overlap case ($s=0$ since $\mathcal{C} \cap \mathcal{R} = \varnothing$) to the perfect one ($s=1$ if $\mathcal{C} = \mathcal{R}$).

\begin{table}[t!]
	\centering
	\caption{Noisy curve fitting. S\o{}rensen-Dice index $s$ and reconstruction error for variable noise levels.}
	\label{tab:QS}
	\begin{tabular}{|r|c|c|c|c|c|}
		Noise level & $1\%$ & $10\%$ & $25\%$ & $50\%$ & $75\%$ \\\hline\hline
		$s$ & $1$ & $1$ & $0.89$ & $0.67$ & $0.39$ \\
		error & 0.21 & 0.23 & 0.27 & 0.75 & 1.25 \\\hline
	\end{tabular}
\end{table}

In Table \ref{tab:QS}, despite the increasing noise level, the reconstruction error is quite stable and only degrades at the highest noise levels. Additionally, when the noise level has a minor impact ($1\%$ and $10\%$), we get $s=1$: the removal process is perfect and exactly all the corrupted labels are effectively removed. When percentages of noise increases ($25\%$, $50\%$), we still have good overlapping measures. The final drop at $75\%$ is coherent with the huge amount of noise (only 1 target out of 4 is not corrupted). 

\begin{table*}[t!]
	\centering
	\begin{tabular}{|r|c|c|c|c|}
		{Method} &  \emph{House} &  \emph{Air} &  \emph{Hydro} &  \emph{Wine} \\ \hline\hline
		{GPR} & \multicolumn{4}{c|}{ {Affine mean, mixture covariance type (linear + squared exponential)}} \\ \hline
		{RR} &  {$\alpha = 0.3$} &  { $\alpha = 0.01$} &  {$\alpha = 0.5$} &  {$\alpha= 1$} \\ \hline
		{$K$-nn} &  {$K = 3$} &  {$K = 3$} &  {$K = 3$} &  {$K = 5$} \\ \hline 
		{NN} &  {$n_h = 3$} &  {$n_h = 5$} &  {$n_h = 7$} &  {$n_h = 6$} \\ \hline
		{SVR} &  {$\epsilon=0.003, C = 10$} &  {$\epsilon=0.005, C = 10$} &  {$\nu=0.003, C = 1$} &  {$\nu=0.01, C = 10$} \\\hline
		{HLR} & \multicolumn{4}{c|}{ {$\lambda = 0.001, \gamma = 0.0001, \Delta \xi = 0.01, T = 3$}} \\ \hline
	\end{tabular}
	\caption{\footnotesize In addition to the parameters $\lambda,\gamma, \Delta\xi$ and $T$ of HLR, we report the parameters/settings of other methods benchmarked on the UCI Machine Learning Repository experiments: the mean and covariance functions used for GPR, the regularizing parameter $\alpha$ for RR, the value of $K$ neighbours considered, the number of neurons $n_h$ in the hidden layer for NN and the $\epsilon/\nu$ choices for SVR as well as the cost function $C$ used. }
	\label{tab:par22}
\end{table*}

\begin{table*}[t!]
	\centering
	\begin{tabular}{|r|c|c|c|c|c|c|c|c|c|c|c|}
		& \multicolumn{3}{c|}{ \emph{House}} & \multicolumn{3}{c|}{ \emph{Air}} & \multicolumn{2}{c|}{ \emph{Hydro}} & \multicolumn{3}{c|}{ \emph{Wine}} \\
		{Methods} &  {\MAE} &  {\MSE} &  {\MRE} &  {\MAE} &  {\MSE} &  {\MRE} &  {\MAE} &  {\MSE} &  {\MAE} &  {\MSE} &  {\MRE} \\\hline\hline
		{GPR} &  {\bf 4.21(3)} &  {\bf 41.00(3)} &  {\bf 0.20(3)} &  {\bf 4.47(2)} &  {\bf 33.84(2)} &  {\bf 0.03(1)} &  {\bf 7.10(2)} &  {\bf 118.3(3)} &  {\bf 0.59(1)}&  {\bf 0.72(1)}&  {\bf 0.107(2)}\\
		{RR} &  {\bf 3.79(1)} &  {\bf 28.73(1)} &  {\bf 16.03(2)} &  {\bf 4.76(3)} &  {\bf 37.61(3)}&  {\bf 3.87(3)}&  {\bf 7.28(3)}&  {\bf 113.9(2)}&  {\bf 0.59(1)}&  {\bf 0.72(1)}&  {\bf 0.106(1)} \\
		{$K$-nn}&  {5.91(6)} &  {64.96(6)} &  {22.64(6)} &  {6.01(5)} &  {65.57(6)} &  {4.89(5)} &  {9.08(6)} &  {267.0(6)} &  {0.61(5)} &  {0.78(5)} &  {0.108(4)} \\
		{NN}&  {5.49(5)} &  {56.97(5)} &  {20.94(5)} &  {6.56(6)} &  {64.69(5)} &  {5.32(6)} &  {8.32(5)} &  {183.2(5)} &  {0.86(6)} &  {1.35(6)} &  {0.161(6)} \\
		{SVR}&  {4.88(4)} &  {51.55(4)} &  {20.72(4)} &  {4.93(4)} &  {38.64(4)} &  {3.99(4)} &  {7.41(4)} &  {143.8(4)} & {\bf 0.59(1)} &  {\bf 0.73(3)} &  {0.109(5)}\\\hline
		{HLR} &  {\bf 4.13(2)} &  {\bf 36.78(2)} &  {\bf 0.15(1)} &  {\bf 4.16(1)} &  {\bf 30.20(1)} &  {\bf 0.04(2)} &  {\bf 6.91(1)} &  {\bf 110.8(1)} & {0.61(4)} &  {0.77(4)} & {\bf 0.107(2)} \\ \hline
	\end{tabular}
	\caption{Comparison of HLR against Gaussian Process Regression, Ridge Regression, $K$ nearest neighbors, neural nets and support vector machine for regression. In bold, top three performing methods. In brackets, the relative ranking. For \emph{Hydro}, since the target variable is sometimes (close to) zero, $\MRE$ metric diverges and therefore was not reported.}\label{tab:ris}
\end{table*}

\emph{Learning with noisy labels.} \quad We want to benchmark HLR in handling noisy annotations adopting the protocol of \cite{NIPSpaper}. Therein, binary classification is performed in the presence of random noise so that some of the positive and negative labels have been randomly flipped with a given probability. Precisely, following \cite{NIPSpaper}, we denote with $\rho_+$ the probability that the label of a positive sample is flipped from $+1$ to $-1$. In a similar manner, $\rho_-$ quantifies the negative instances whose label is wrongly assigned to be $-1$. In \cite{NIPSpaper}, such problem is stated under a theoretical perspective, formulating some bounds for the generalization error and the empirical risk, as to guarantee the feasibility of the learning task even in such an extreme situation. Although interesting per se, such arguments are out of the scope of our work, where, instead, we compared HLR with the two methods proposed by \cite{NIPSpaper}: a surrogate logarithmic loss function ($\widetilde{ \ell }_{\rm log}$) and a variant of support vector machine algorithm, where the cost parameter is adapted depending on the training labels (C-SVM). In \cite{NIPSpaper}, $\widetilde{ \ell }_{\rm log}$ and C-SVM were shown to outperform other methods devised for the identical task: the max-margin perceptron algorithm (PAM) \cite{PAM}, Gaussian herding (NHERD) \cite{NHERD} and random projection classifier (RP) \cite{RP}. All the aforementioned methods are compared with HLR where, as usually done for binary decision boundaries, we exploit the sign of the learnt function $h^\star$ (see Section \ref{ssez:HLR}) to perform classification. To ensure a fair comparison, we reproduce the same experimental protocol (Gunnar Raetsch's training/testing splits and data pre-processing for \emph{Breast Cancer}, \emph{Diabetes}, \emph{Thyroid}, \emph{German}, \emph{Heart}, \emph{Image} benchmark datasets\footnote{\url{http://theoval.cmp.uea.ac.uk/matlab}}) and we compute the testing accuracy with respect to the clean distribution of labels \cite{NIPSpaper}.

From the experimental results reported in Table \ref{tab:fff}, HLR scored a strong performance. Indeed, despite some modest classification results on Thyroid an Diabetes datasets, HLR is able to beat the considered competitors, obtaining the best classification accuracy in the remaining 4 out of 6 ones (\emph{Breast Cancer}, \emph{German}, \emph{Heart} and \emph{Image}). Interestingly, this happens in both low and high noise levels: for instance, when $\rho_+ = \rho_- = 0.2$ on \emph{Breast Cancer} and when $\rho_+ = \rho_- = 0.4$ on \emph{Image}, respectively. 

The results presented in this Section attest the active-learning HLR component to be able to effectively detect the presence of outlier data while, at the same time, guaranteeing an effective learning of the regression model.

\subsection{HLR for classical regression applications}\label{sez:altri}

To compare the effectiveness of HLR in learning the regressor map, in this Section, we benchmark on four datasets from the UCI Machine Learning Repository\footnote{ \url{https://archive.ics.uci.edu/ml/datasets}}, we will focus on house pricing estimation (Boston Housing -- \emph{House}), physical simulations (AirFoil Self-Noise -- \emph{Air} and Yatch Hydrodynamics -- \emph{Hydro}) and agronomic quality control (\emph{Wine}). We will briefly discribe each of them. 

\begin{figure*}[t!]		
	\centering
	\subfigure[\emph{House} \label{s:house}]{\includegraphics[keepaspectratio, width=.25\textwidth]{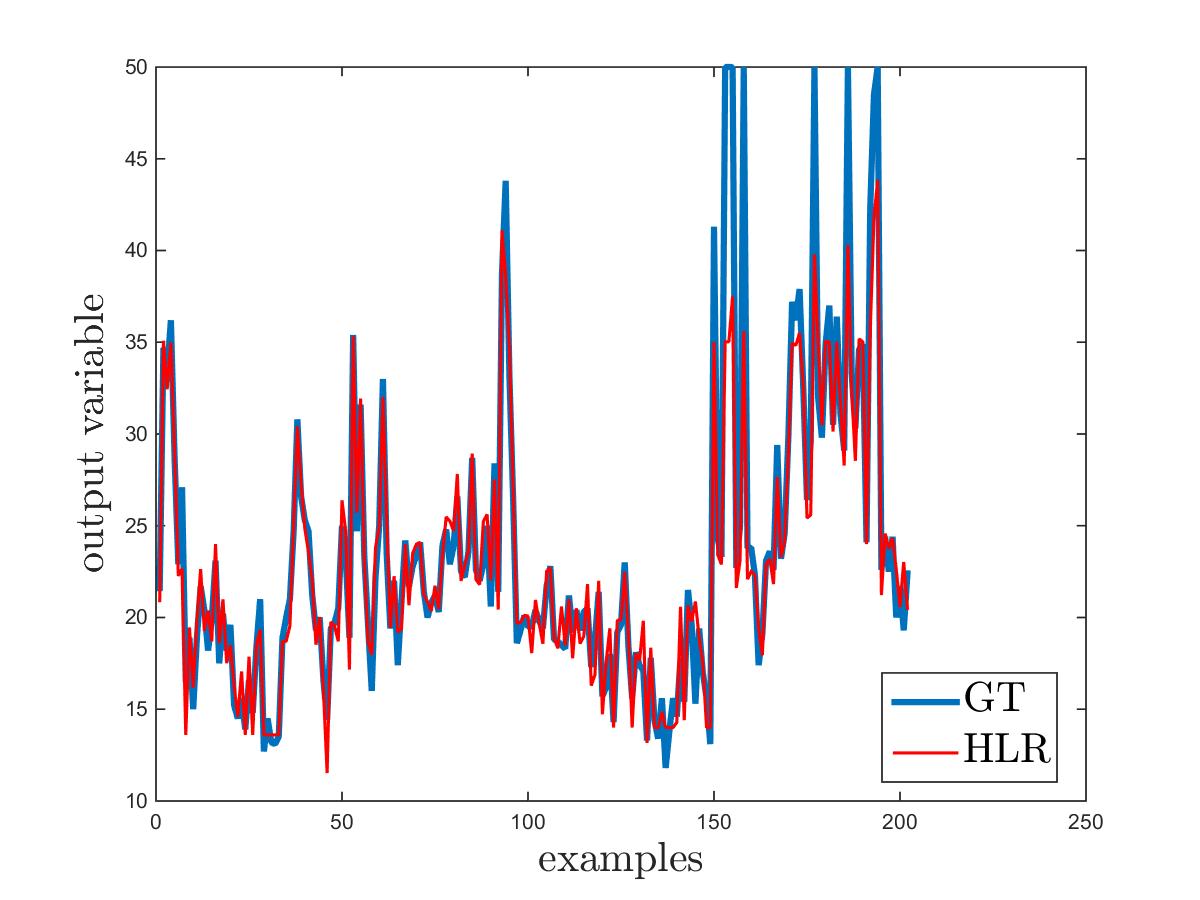}}%
	\subfigure[\emph{Air} \label{s:air}]{\includegraphics[keepaspectratio, width=.25\textwidth]{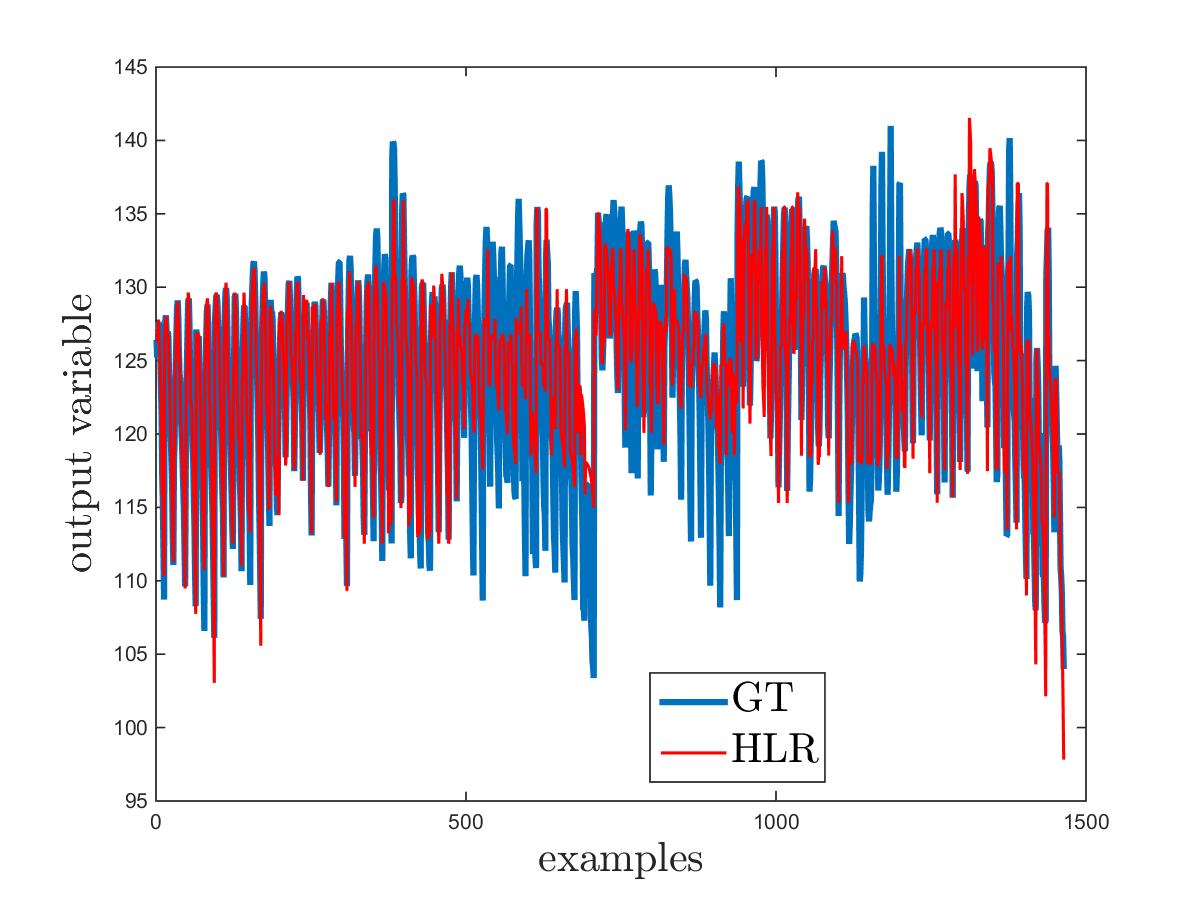}}%
	\subfigure[\emph{Hydro} \label{s:idro}]{\includegraphics[keepaspectratio, width=.25\textwidth]{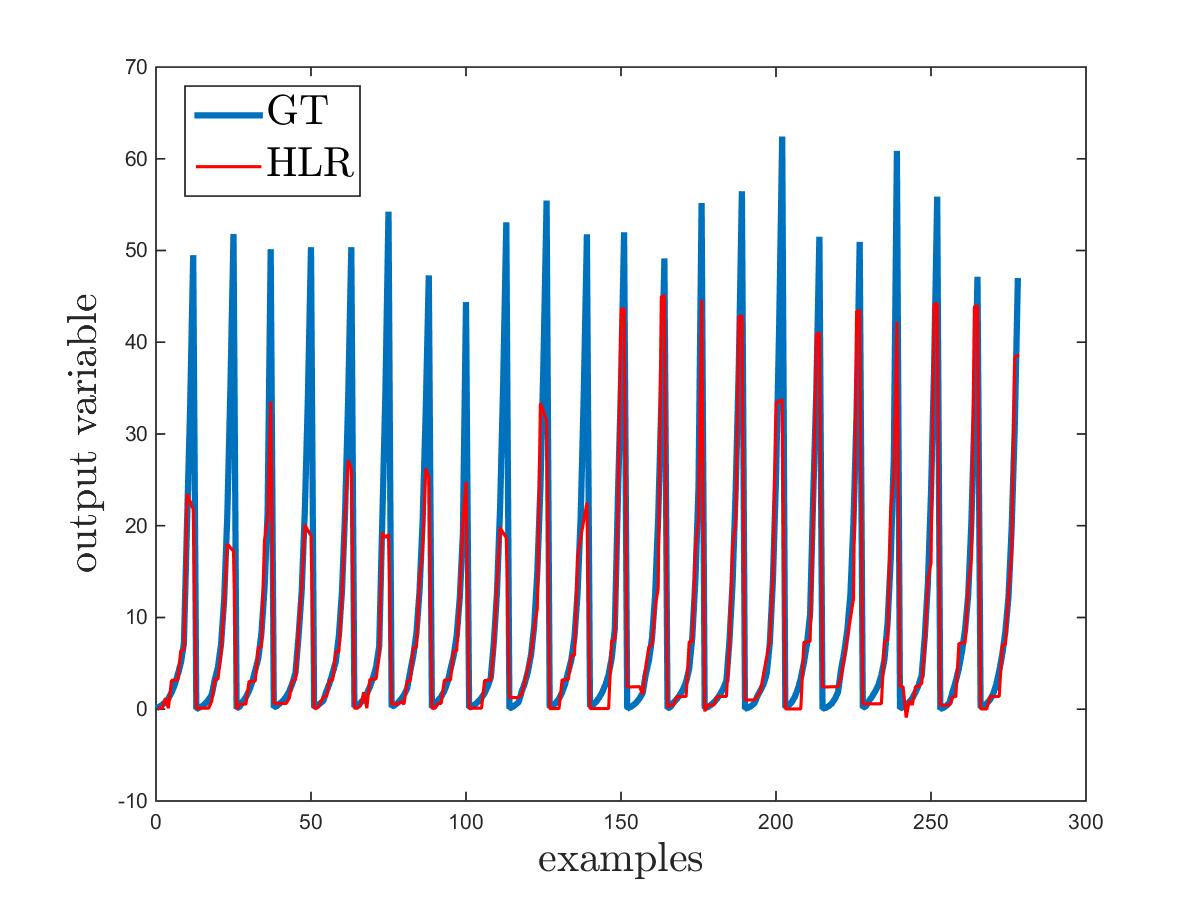}}%
	\subfigure[\emph{Wine} \label{s:wine}]{\includegraphics[keepaspectratio, width=.25\textwidth]{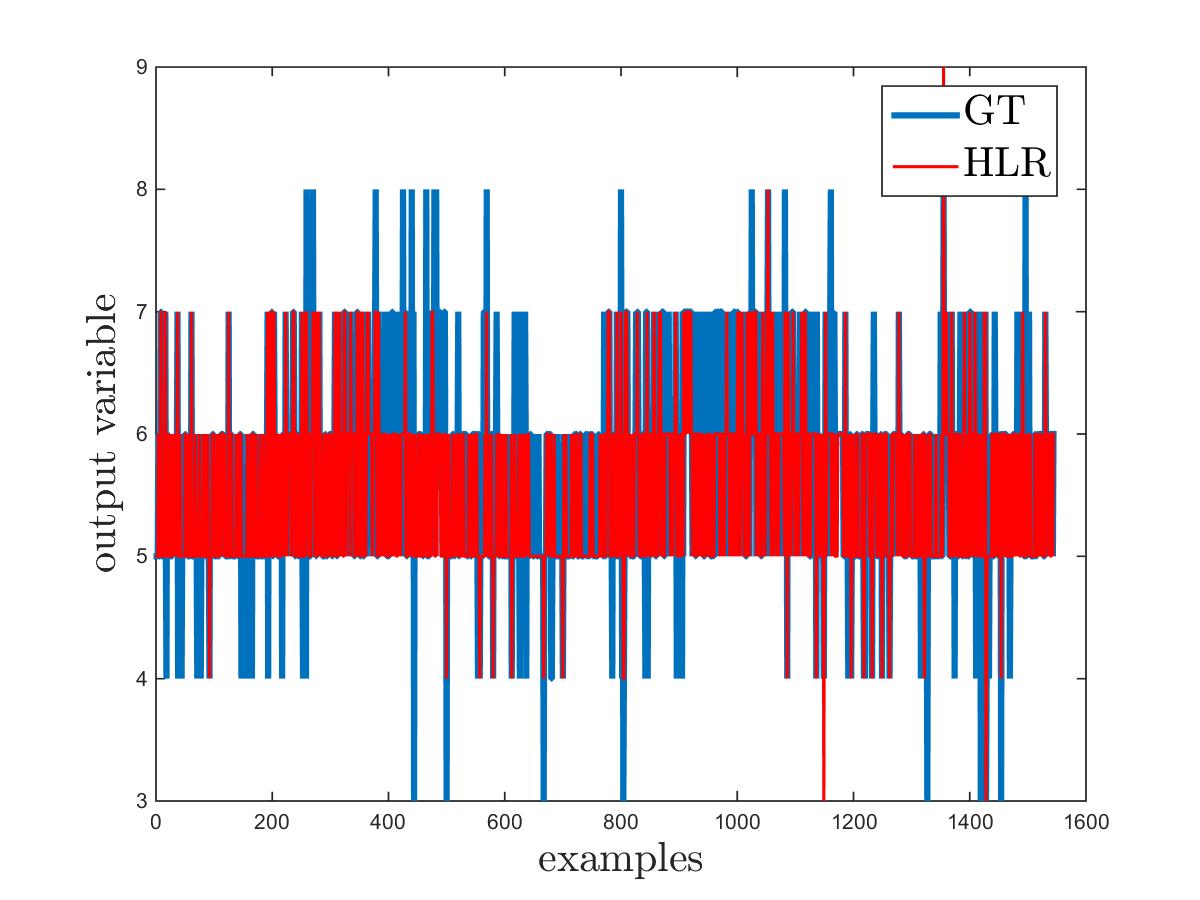}} 
	\caption{Ground truth (GT) compared with HLR prediction for UCI Machine Learning Repository datasets. Best viewed in colors.}\label{fig:altritask}
\end{figure*}

\emph{House} datasets predicts housing values in Boston suburbs. The dataset consists in 506 examples and 13 feature components which are either discrete (average number of rooms), binary (whether or not tracting bounds of Charles river) or continuous (pupil-teacher ratio by town). \emph{Air} datasets address the problem of physical simulations. It is provided by NASA and shows 1503 aerodynamic and acoustic acquisitions of two and three-dimensional air foil blade sections. A 6-dimensional feature vector encodes different size and thickness for blades, various wind tunnel speeds and angles of attack. The output variable is the sound pressure level measured in decibel. \emph{Hydro} predicts the resistance of sailing yachts at the initial design stage, estimating the required propulsive power. Inputs provides hull dimensions an boat velocity (6 dimensional features, 308 instances). The output variable is the residuary resistance per unitary displacement. \emph{Wine} dataset consists in 11-dimensional 1599 input instances (we only focused on red wine). The goal is predicting the ratings, given by a crew of sommeliers, as function of pH and alcohol/sulphates concentrations.

Over the aforementioned datasets, we compare Huber loss regression (HLR) against Gaussian process regression (GPR), ridge regression (RR),  $K$ nearest neighbours ($K$-nn), one-hidden-layer neural network (NN) and linear support vector machine for regression (SVR). For each method, the parameters setting are obtained after cross validation (see Table \ref{tab:par22}). For a fair comparison, we split each dataset in five equispaced folds and performing a leave-one-fold-out testing strategy. To give a comprehensive results on each datasets, we averaged the errors on each fold using one out of those following metrics\begin{align}
\mbox{mean absolute error} \qquad  &\MAE= \frac{1}{n} \sum_{i=1}^n |y_i - \widehat{y}_i|, \\
\mbox{mean squared error} \qquad &\MSE = \frac{1}{n} \sum_{i=1}^n (y_i - \widehat{y}_i)^2, \\
\mbox{mean relative error} \qquad &\MRE= \frac{1}{n} \sum_{i=1}^n \frac{|y_i - \widehat{y}_i|}{y_i},
\end{align}
where $y_1,\dots,y_n$ are the (true) testing target variables and $\widehat{y}_1,\dots,\widehat{y}_n$ the corresponding estimated predictions. 

Qualitative and quantitative analysis has been reported in Table \ref{tab:ris} and Figure \ref{fig:altritask}, respectively. Globally, HLR shows remarkable performances since outstanding the other methods in 5 cases out of 12. Those performances are also remarkable since they have been obtained with a fixed set of parameters, confirming the ductility of HLR (see Table \ref{tab:par22}). Indeed, despite GPR and RR scored comparable performance with respect to HLR, the regularizing parameter of RR has to be tuned and the parameters of the GPR has to be learnt in a maximum likelihood sense (mean function plus covariance kernel). 

From this analysis, the low scored errors and the fixed parameter configuration make HLR outperforming many state-of-the-art approaches for scalar regression tasks.

\subsection{HLR for crowd counting application}\label{sez:CC}

As a final test bed of our proposed framework, we address the crowd counting application, namely estimating the number of people in a real world environment using video data. Crowd counting can be rephrased in learning a regression map from frame-specific features to the amount of people whereby \cite{Gong}. Three benchmark datasets have been used to test the performances of our Huber loss regression. They are \emph{MALL} \cite{Chen:BMVC12}, \emph{UCSD} \cite{Chan:CVPR2008} and \emph{PETS 2009} \cite{Chan:2009}.

\begin{figure}[H]
	\centering
	\includegraphics[width=.35\columnwidth]{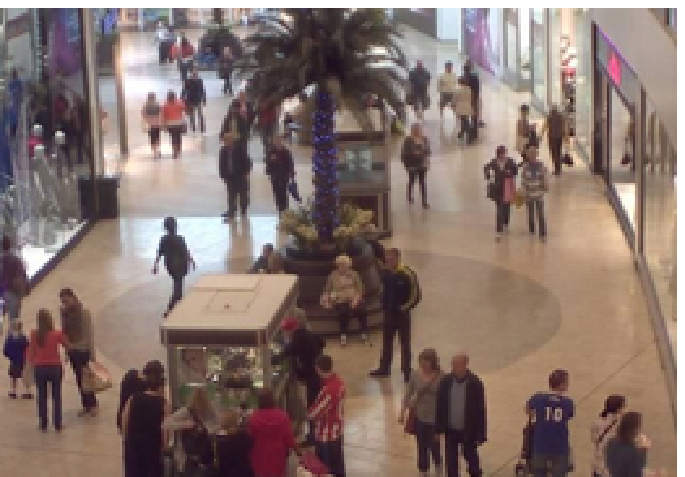}%
	\includegraphics[width=.3\columnwidth]{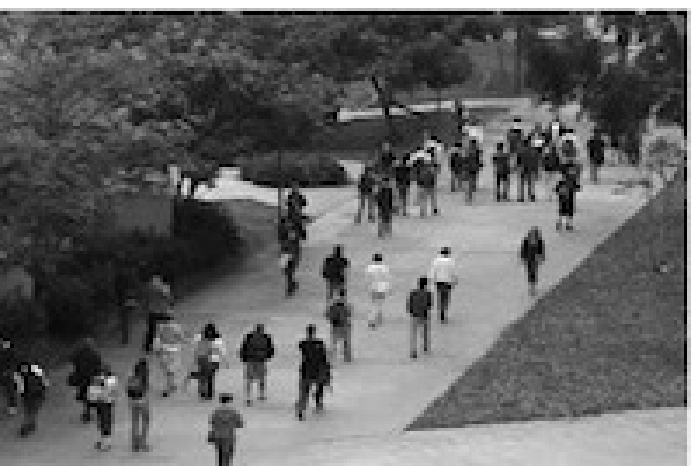}%
	\includegraphics[width=.35\columnwidth]{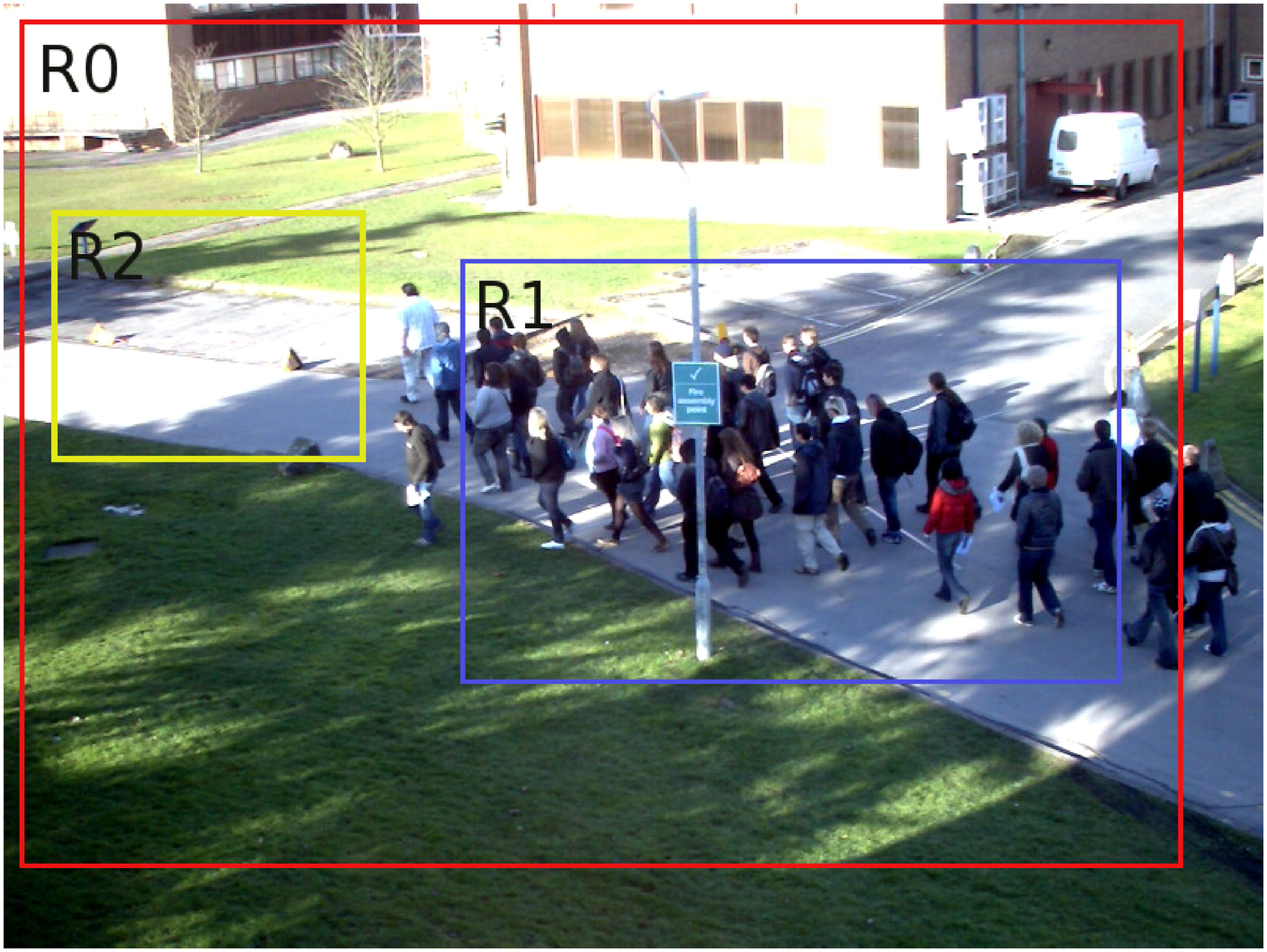}
	\caption{An exemplar frame from \emph{MALL} (left), \emph{UCSD} (center) and \emph{PETS 2009} (right) datasets. For the latter, R0, R1 and R2 regions are represented.}\label{fig:data}
\end{figure}

\begin{table}[H]
	\centering
	\begin{tabular}{|r|c|l|}
		\multicolumn{1}{|c}{Test} & \multicolumn{1}{|c|}{Region(s)} & \multicolumn{1}{c|}{Train} \\ \hline\hline
		$13$-$57$ ($221$) & R0, R1, R2 & $13$-$59$, $14$-$03$ ($1308$) \\\hline
		$13$-$59$ ($241$) & R0, R1, R2 & $13$-$57$, $14$-$03$ ($1268$) \\\hline
		$14$-$06$ ($201$) & R1, R2 & $13$-$57$, $13$-$59$, $14$-$03$ ($1750$) \\\hline
		$14$-$17$ ($91$) & R1 & $13$-$57$, $13$-$59$, $14$-$03$ ($1750$)\\\hline
	\end{tabular}
	\caption{For each \emph{PETS 2009} sequence, the region(s) of interest and the training/testing sets used by \cite{Chan:2009} are provided. Also, in brackets, we report the number of frames involved).}\label{tab:1}
\end{table}

\begin{figure*}[t!]
	\centering
	\subfigure[\emph{UCSD} Table \ref{tab:ryan} \label{g:1}]{\includegraphics[keepaspectratio, width=.25\textwidth]{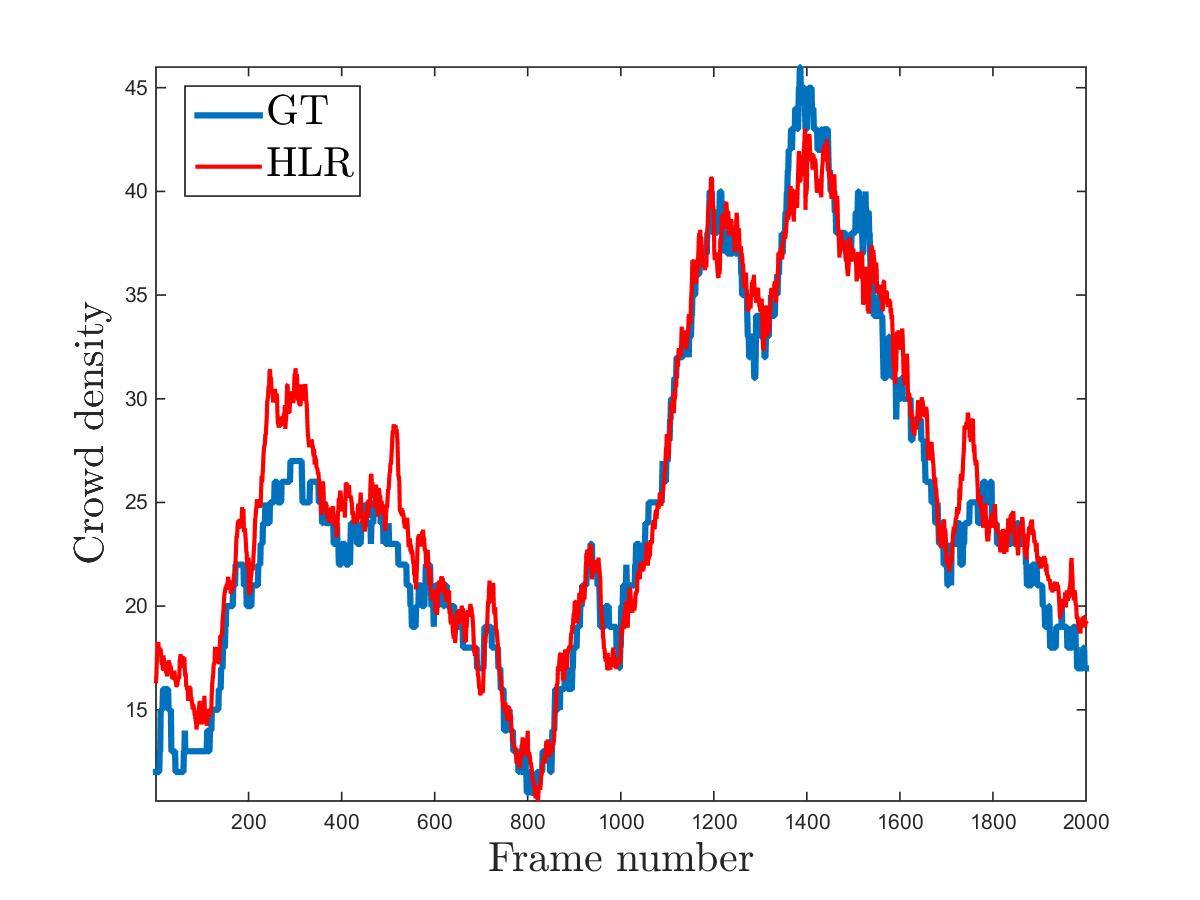}}%
	\subfigure[\emph{MALL} Table \ref{tab:ryan} \label{g:2}]{\includegraphics[keepaspectratio, width=.25\textwidth]{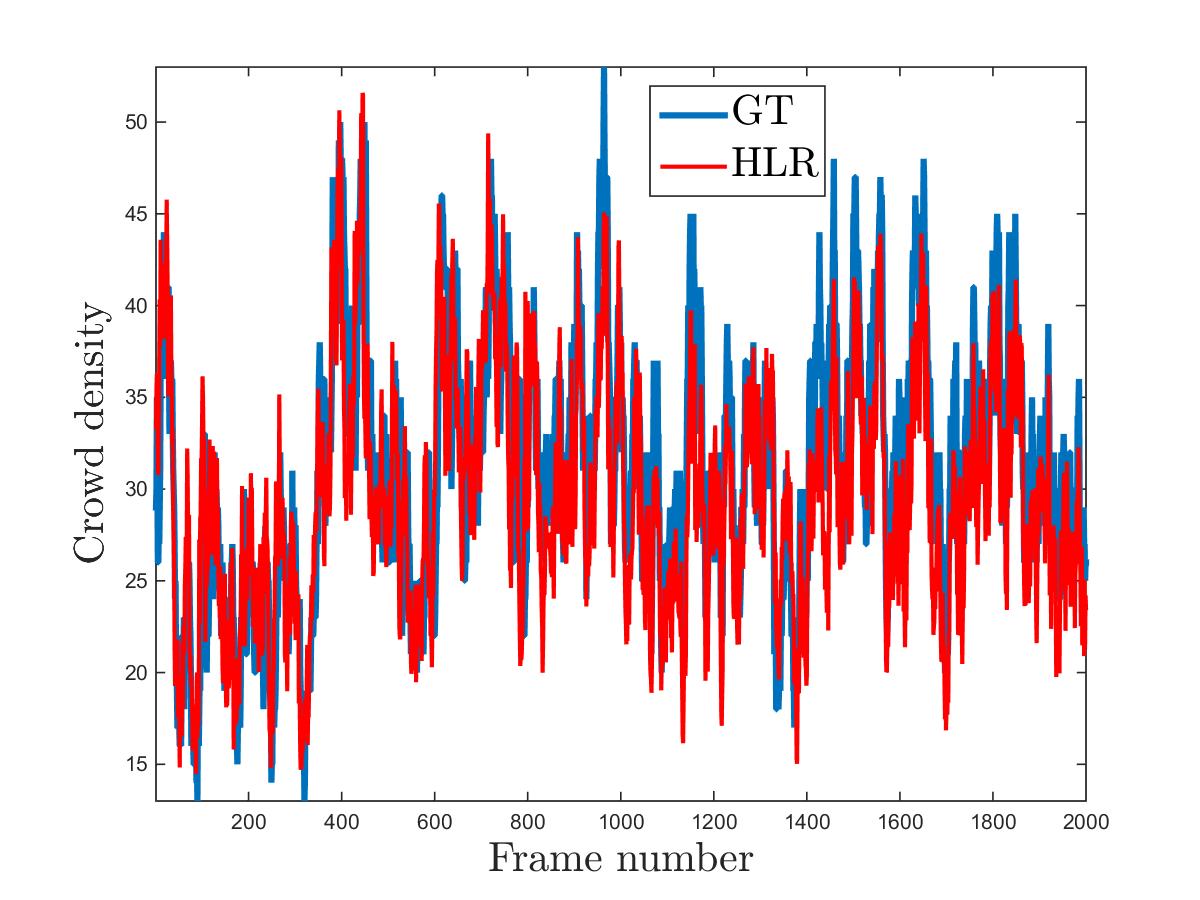}}%
	\subfigure[\emph{UCSD} Tables \ref{tab:semi} \& \ref{tab:multi} \label{g:3}]{\includegraphics[keepaspectratio, width=.25\textwidth]{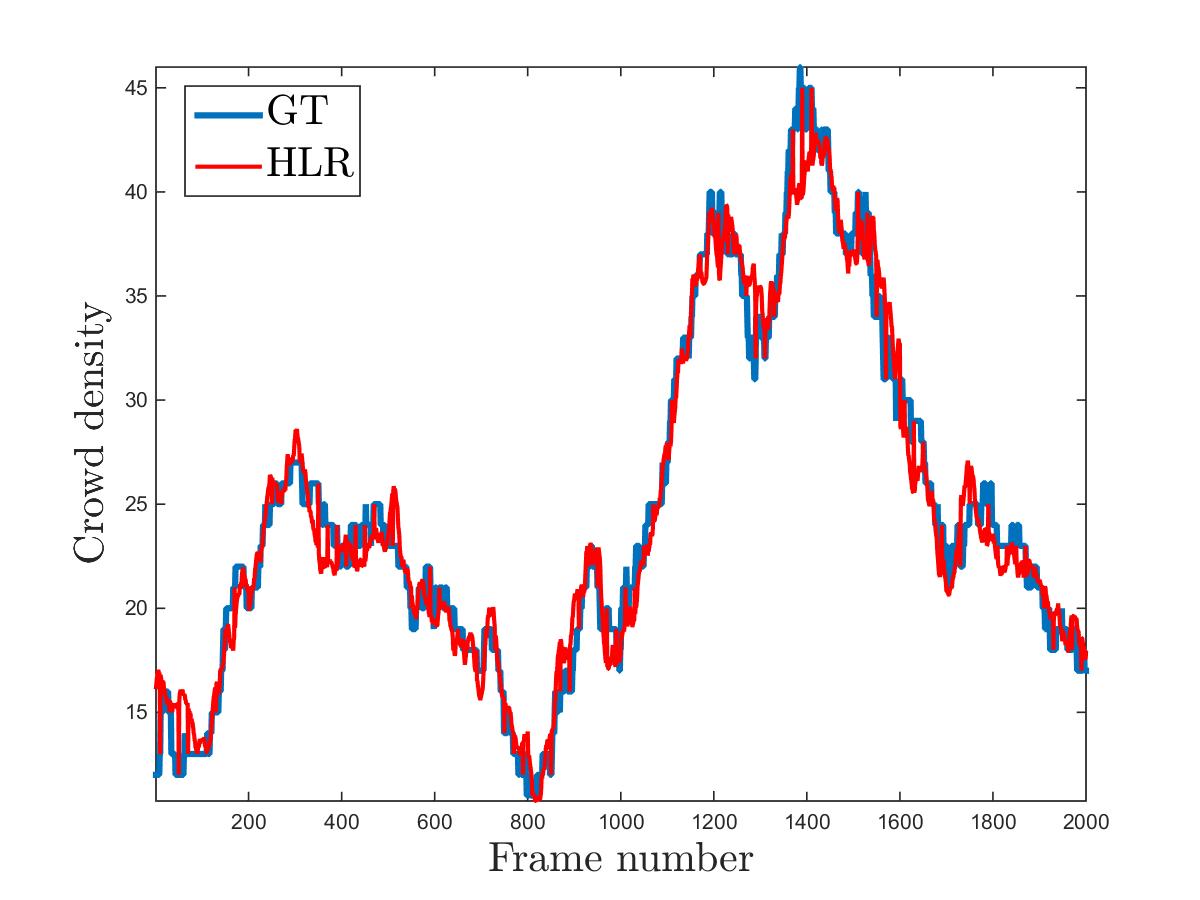}}%
	\subfigure[\emph{MALL} Tables \ref{tab:semi} \& \ref{tab:multi} \label{g:4}]{\includegraphics[keepaspectratio, width=.25\textwidth]{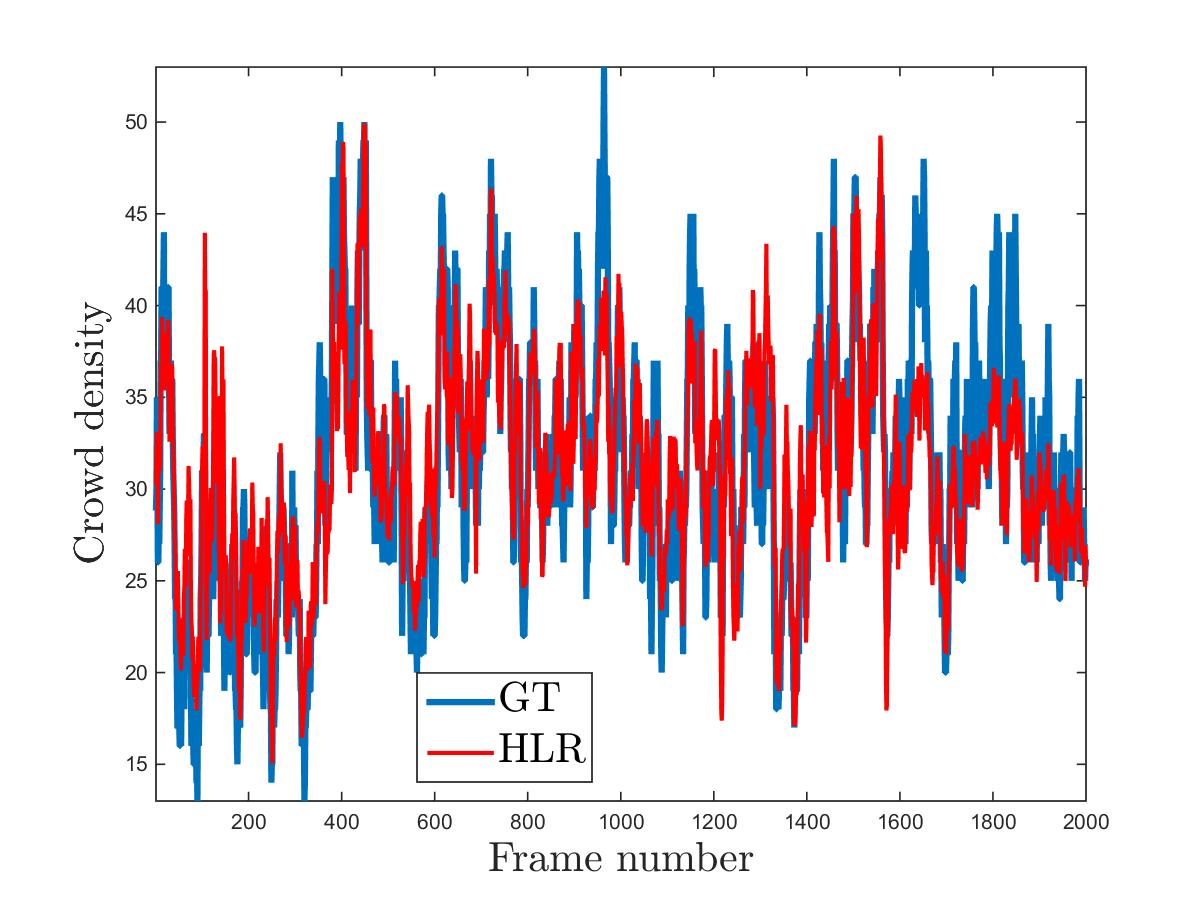}}\\
	\subfigure[$13-57$ R0 left\label{g:e}]{\includegraphics[keepaspectratio, width=.2\textwidth]{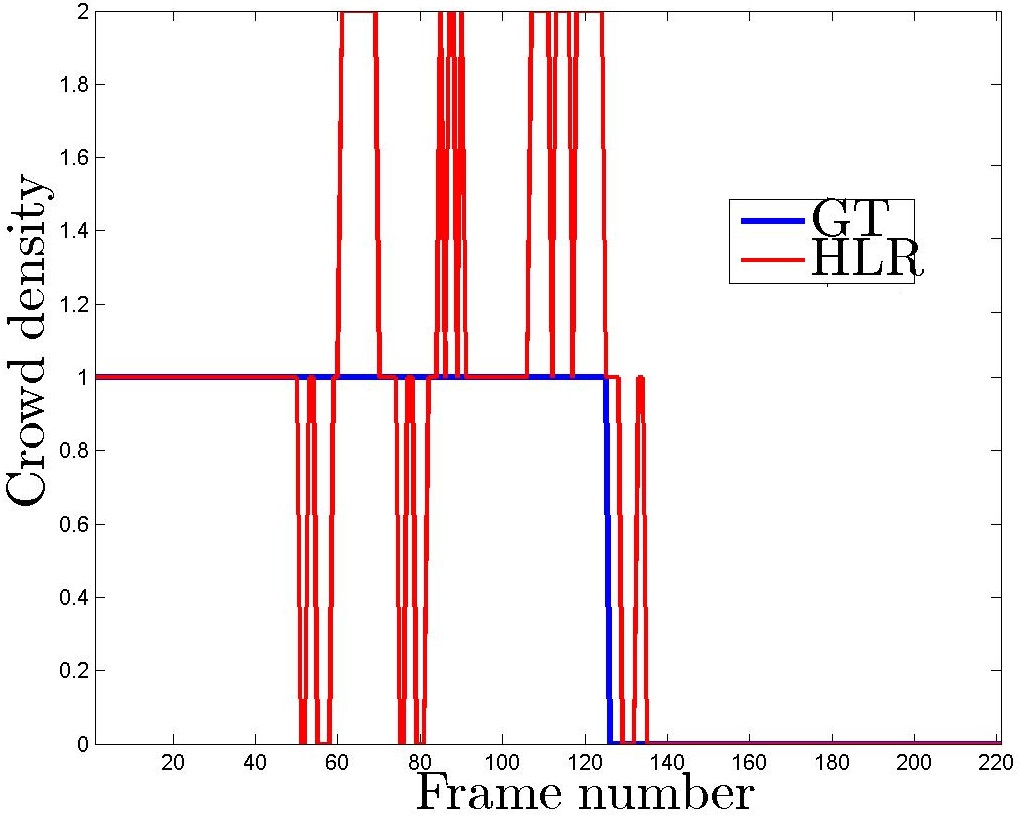}}%
	\subfigure[$13-57$ R0 right \label{g:f}]{\includegraphics[keepaspectratio, width=.2\textwidth]{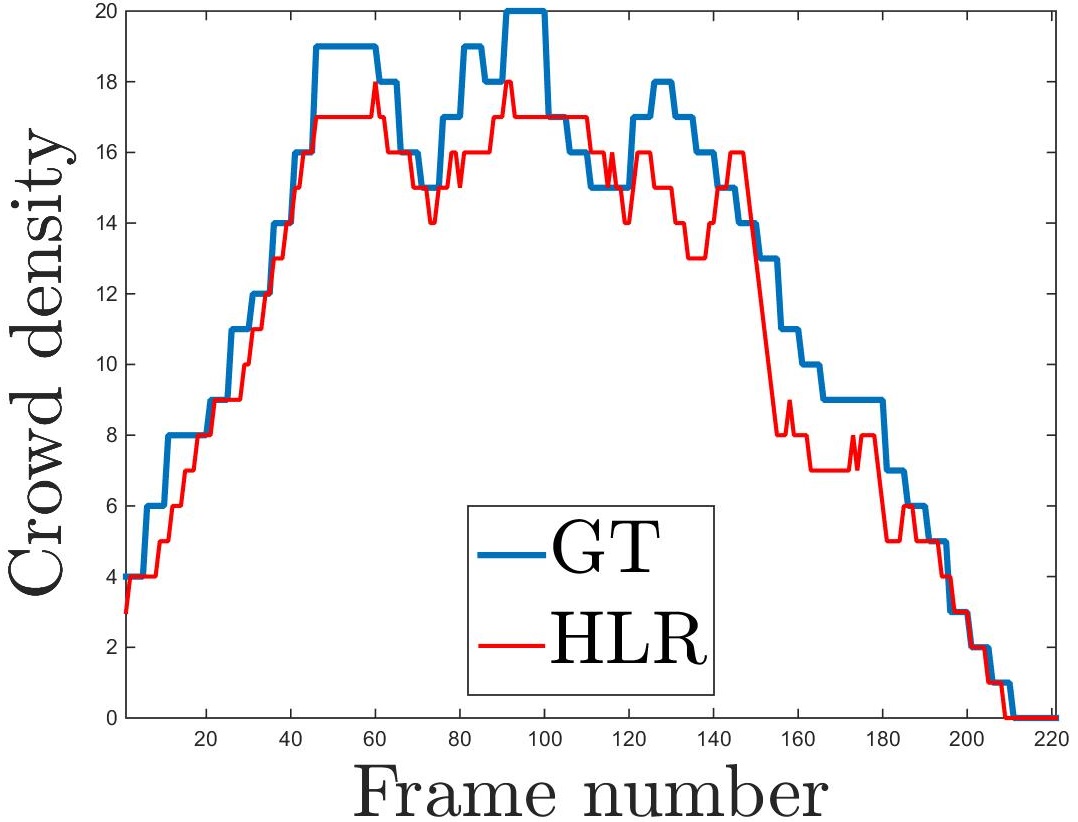}}%
	\subfigure[$13-57$ R0 total\label{g:g}]{\includegraphics[keepaspectratio, width=.2\textwidth]{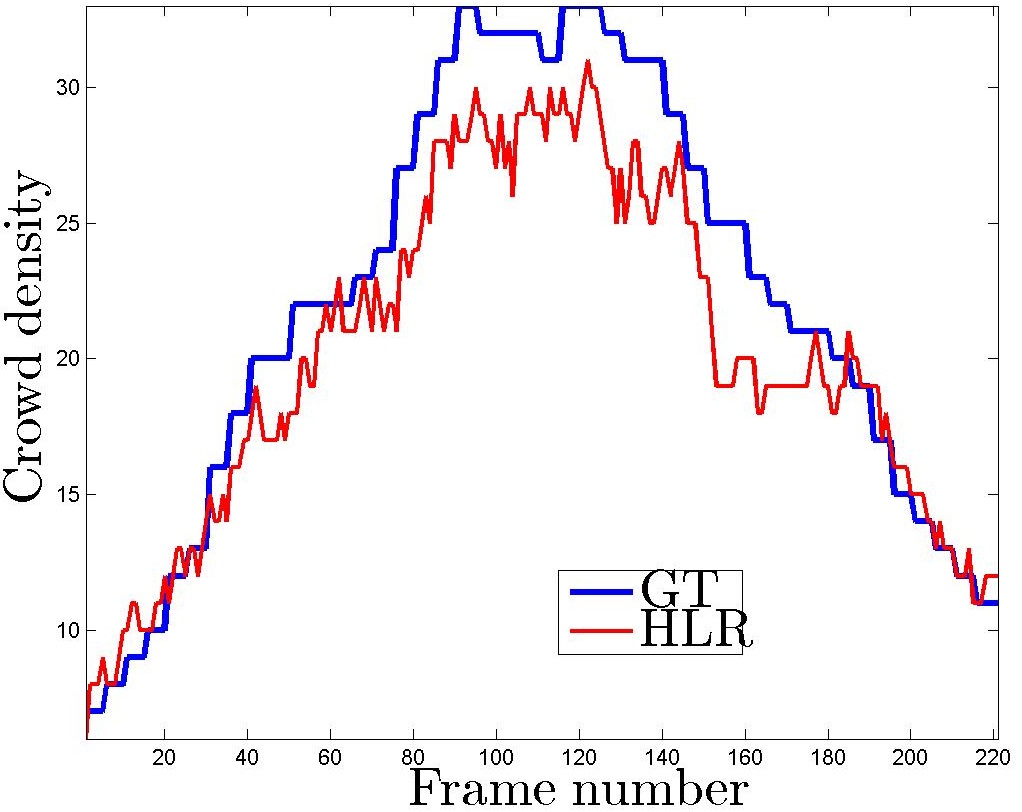}}%
	\subfigure[$13-57$ R1 left \label{g:h}]{\includegraphics[keepaspectratio, width=.2\textwidth]{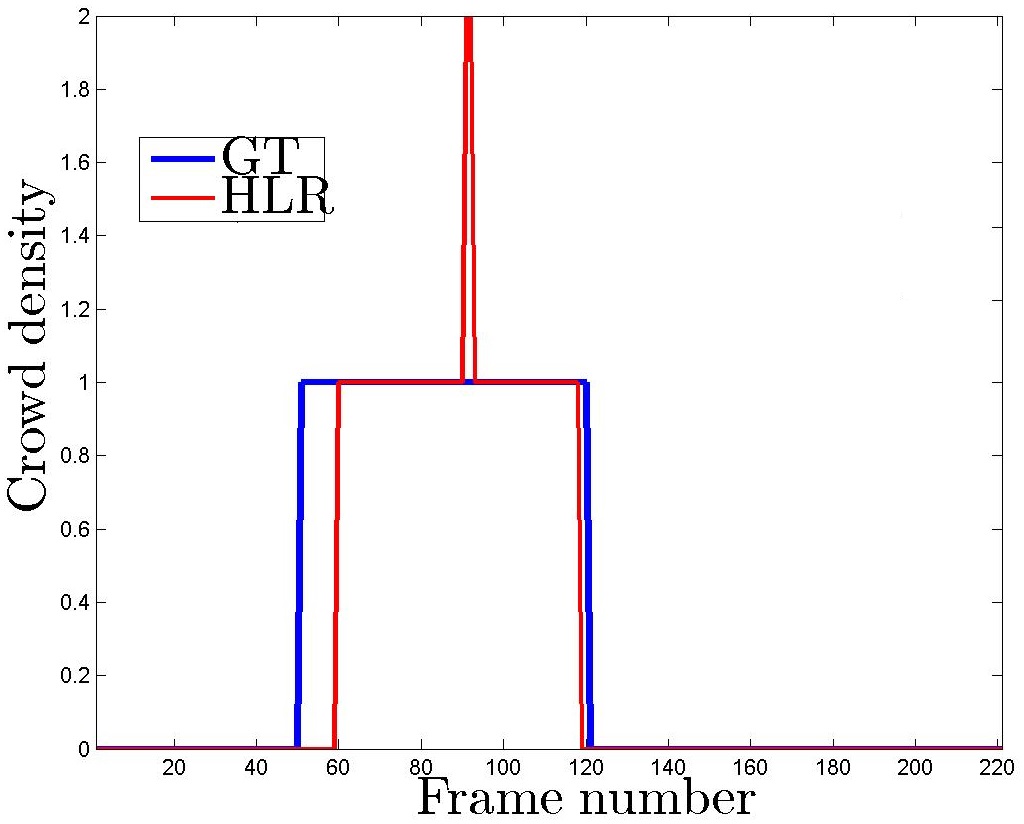}}%
	\subfigure[$13-57$ R1 right \label{g:i}]{\includegraphics[keepaspectratio, width=.2\textwidth]{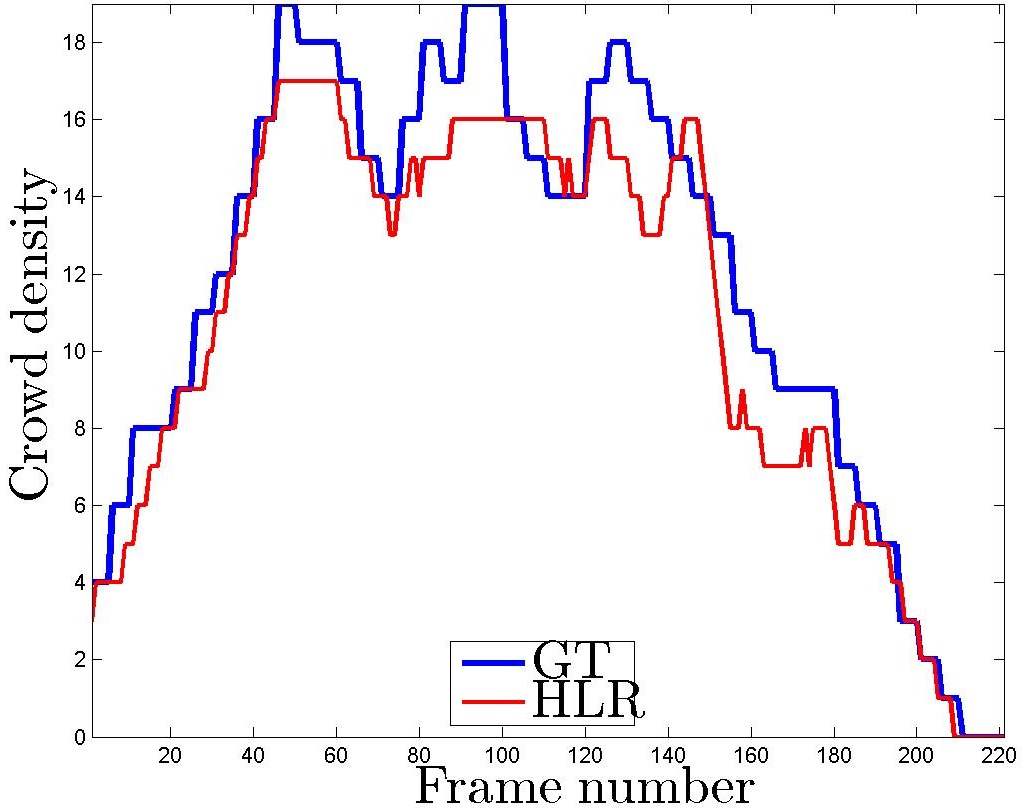}}\\
	\subfigure[$13-57$ R1 total\label{g:j}]{\includegraphics[keepaspectratio, width=.2\textwidth]{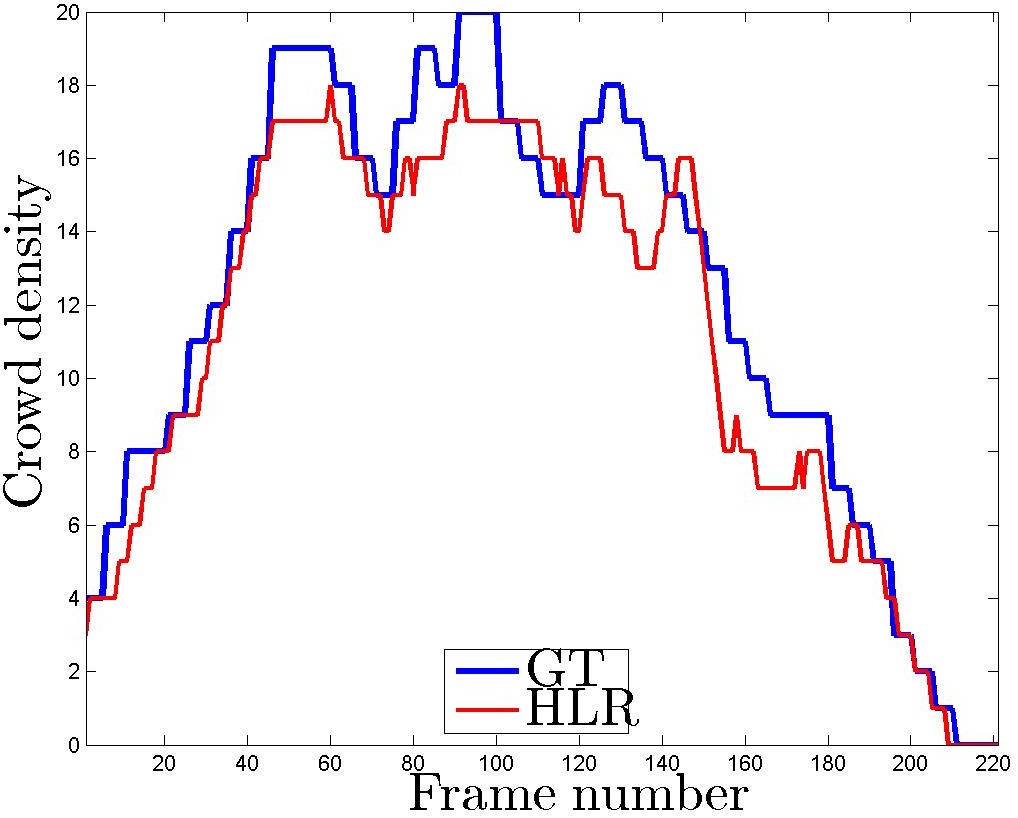}}%
	\subfigure[$13-57$ R2 left \label{g:k}]{\includegraphics[keepaspectratio, width=.2\textwidth]{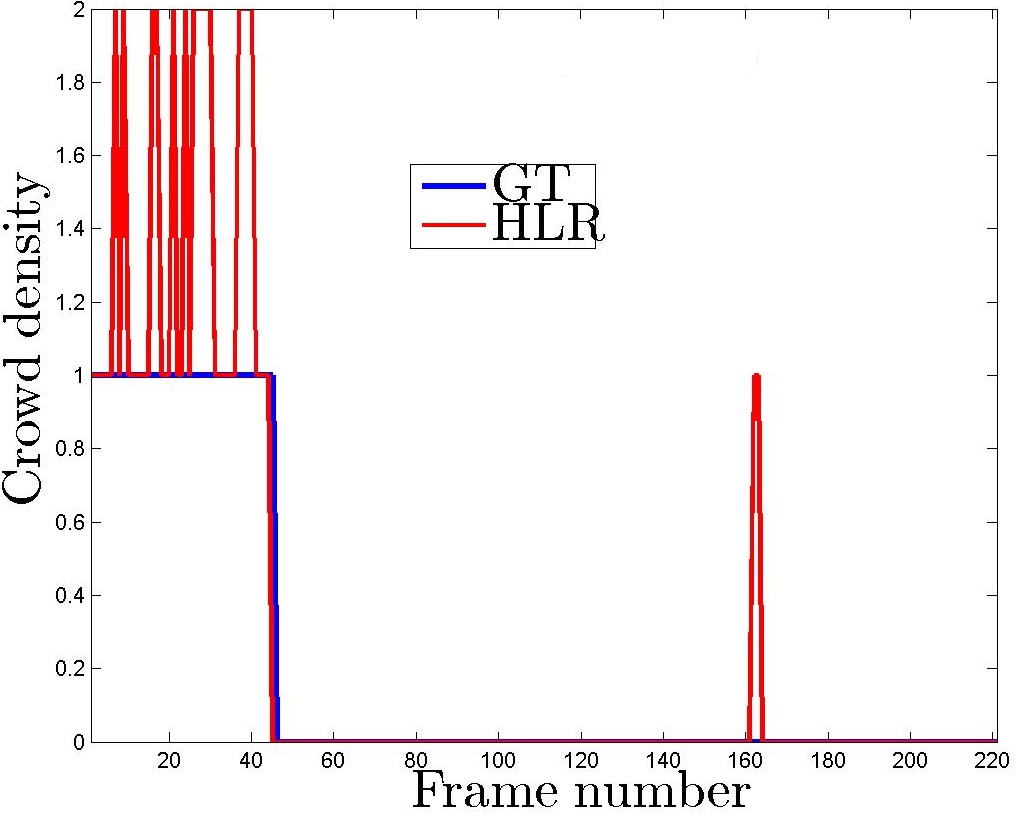}}%
	\subfigure[$13-57$ R2 right \label{g:l}]{\includegraphics[keepaspectratio, width=.2\textwidth]{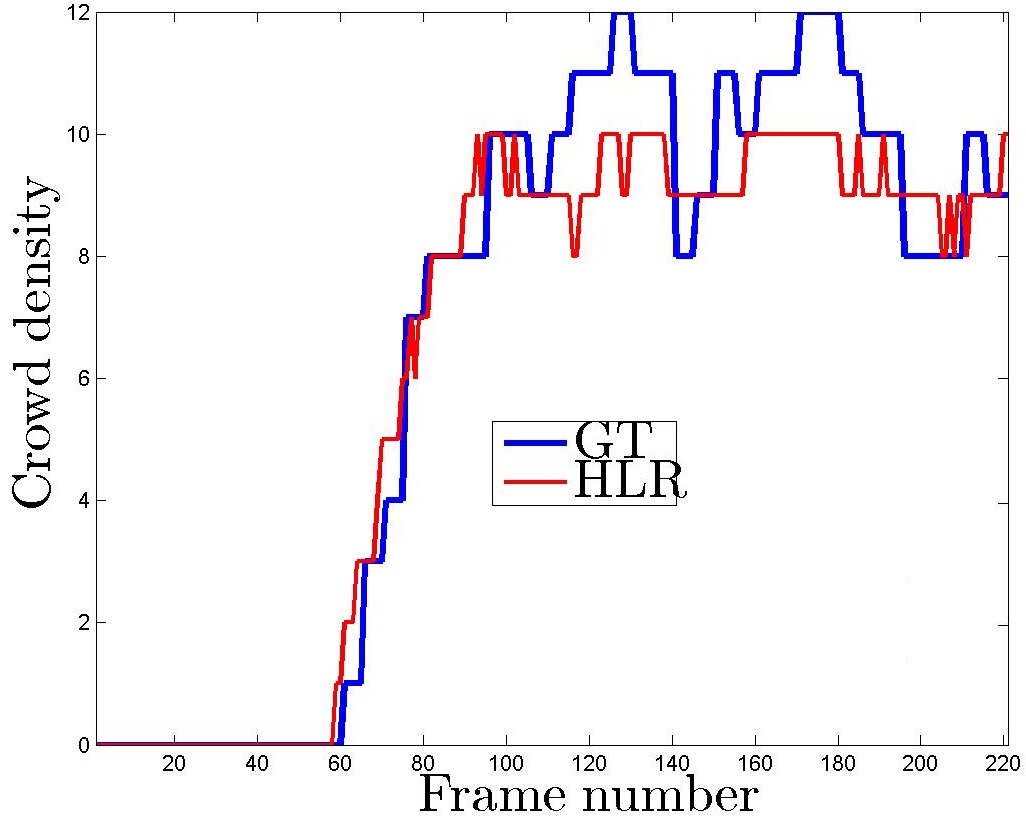}}%
	\subfigure[$13-57$ R2 total\label{g:m}]{\includegraphics[keepaspectratio, width=.2\textwidth]{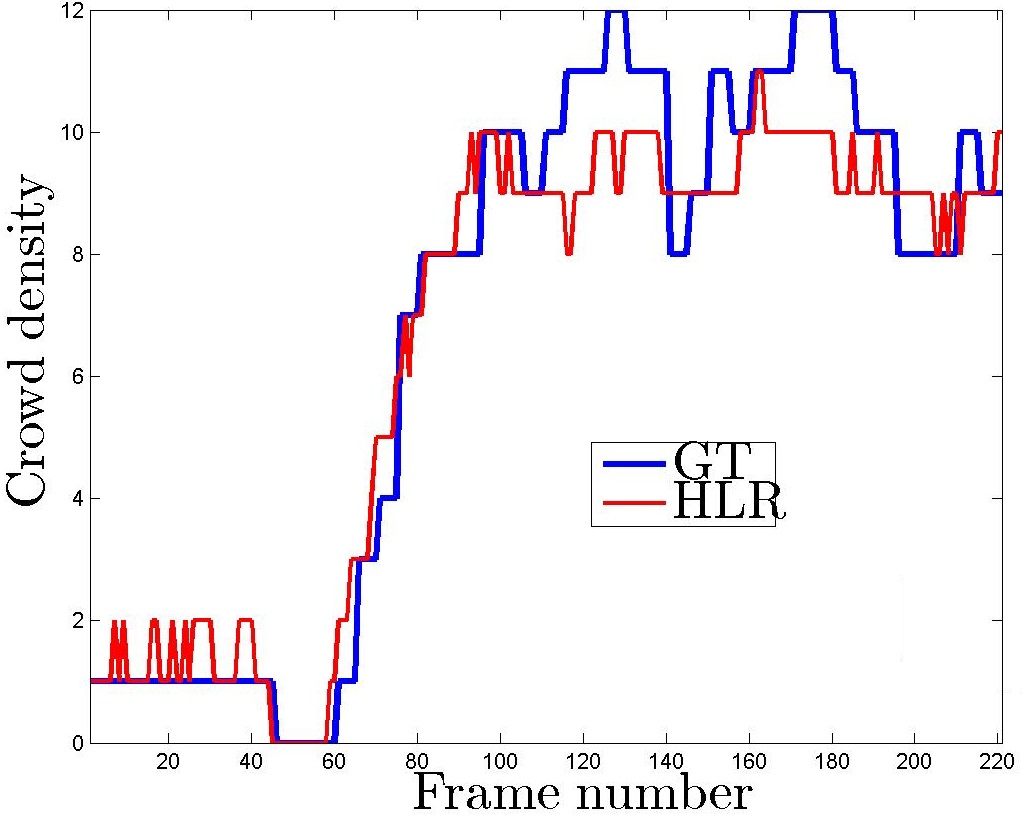}}%
	\subfigure[$13-59$ R0 left \label{g:n}]{\includegraphics[keepaspectratio, width=.2\textwidth]{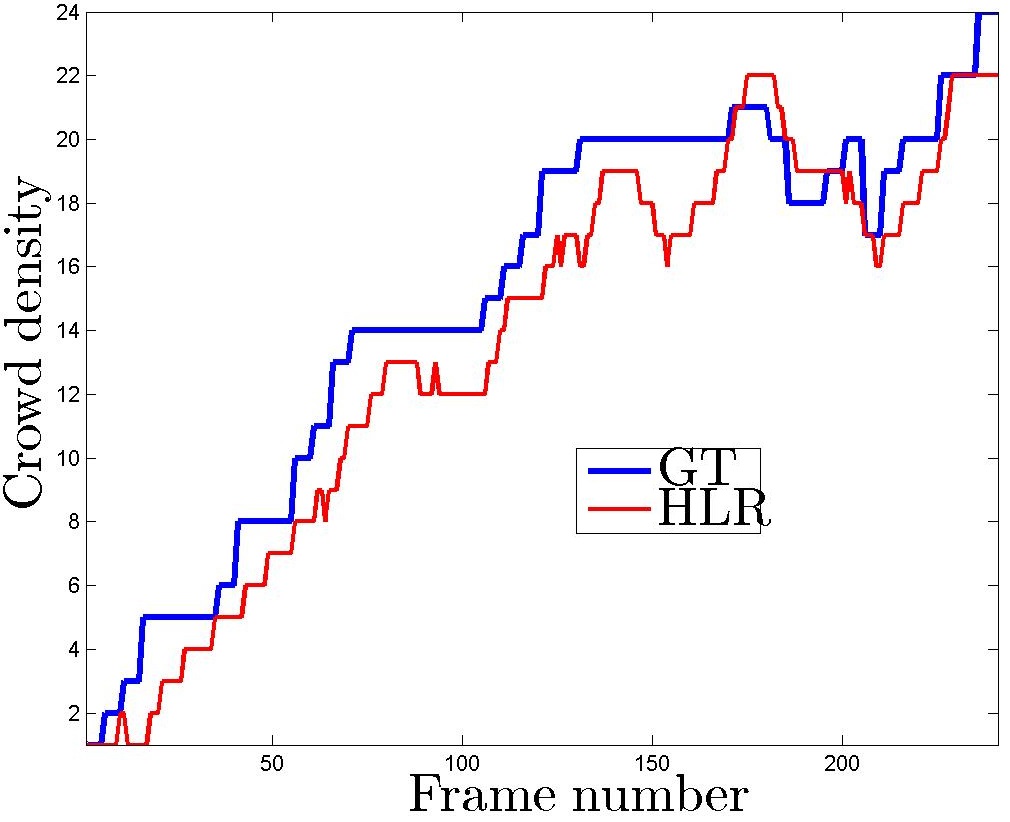}}\\
	\subfigure[$13-59$ R0 right \label{g:o}]{\includegraphics[keepaspectratio, width=.2\textwidth]{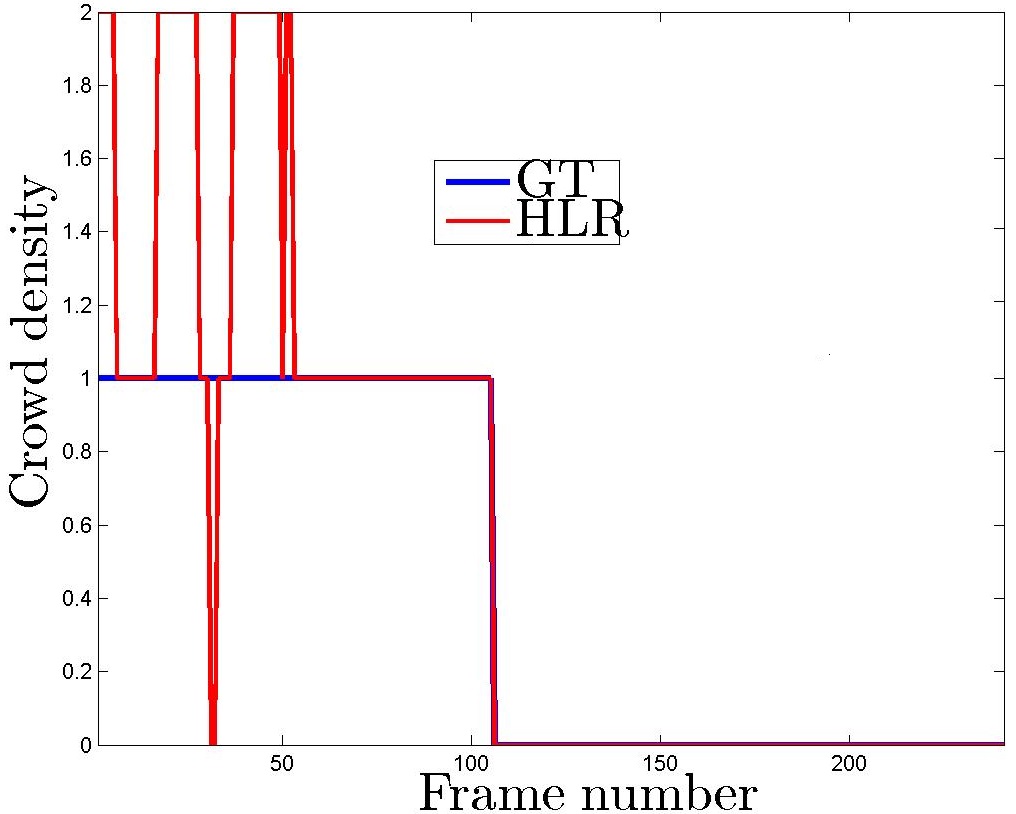}}%
	\subfigure[$13-59$ R0 total\label{g:p}]{\includegraphics[keepaspectratio, width=.2\textwidth]{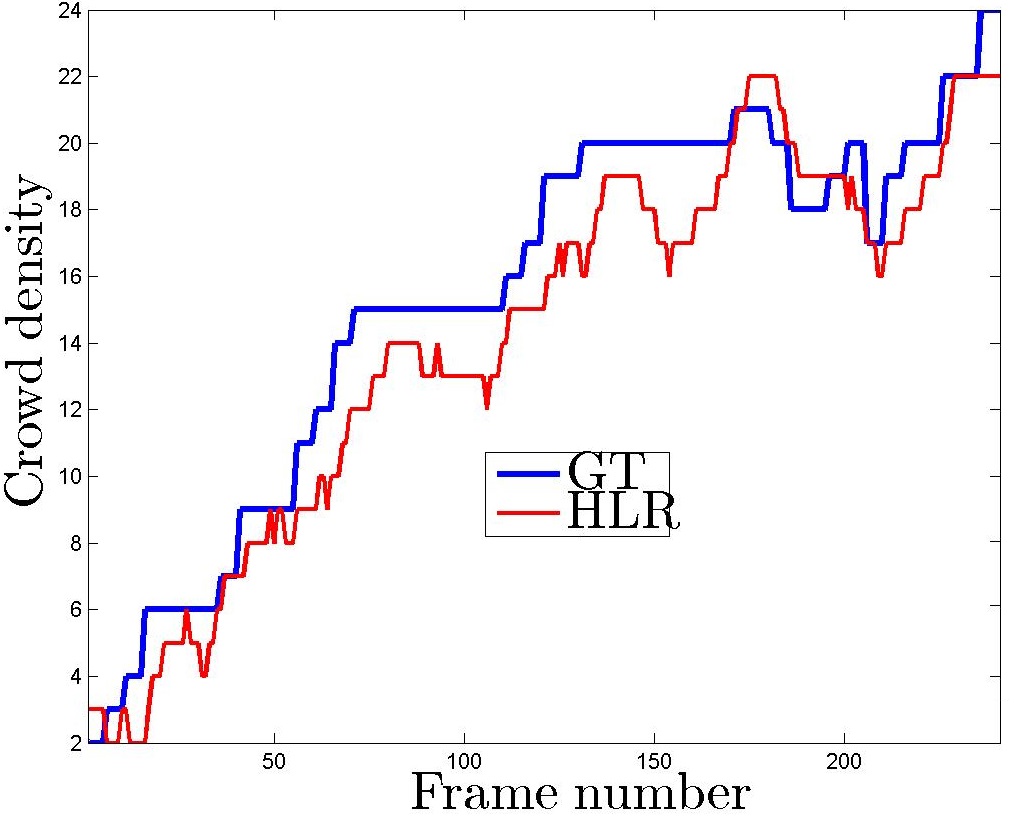}}%
	\subfigure[$13-59$ R1 left\label{g:q}]{\includegraphics[keepaspectratio, width=.2\textwidth]{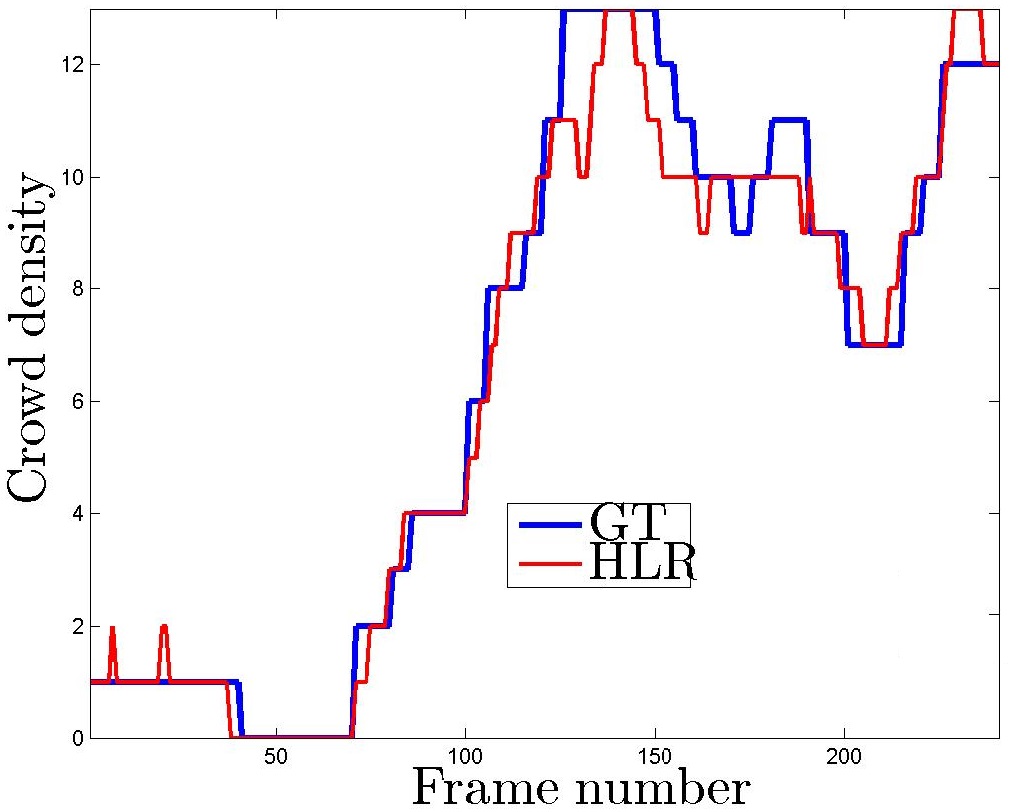}}%
	\subfigure[$13-59$ R1 right \label{g:r}]{\includegraphics[keepaspectratio, width=.2\textwidth]{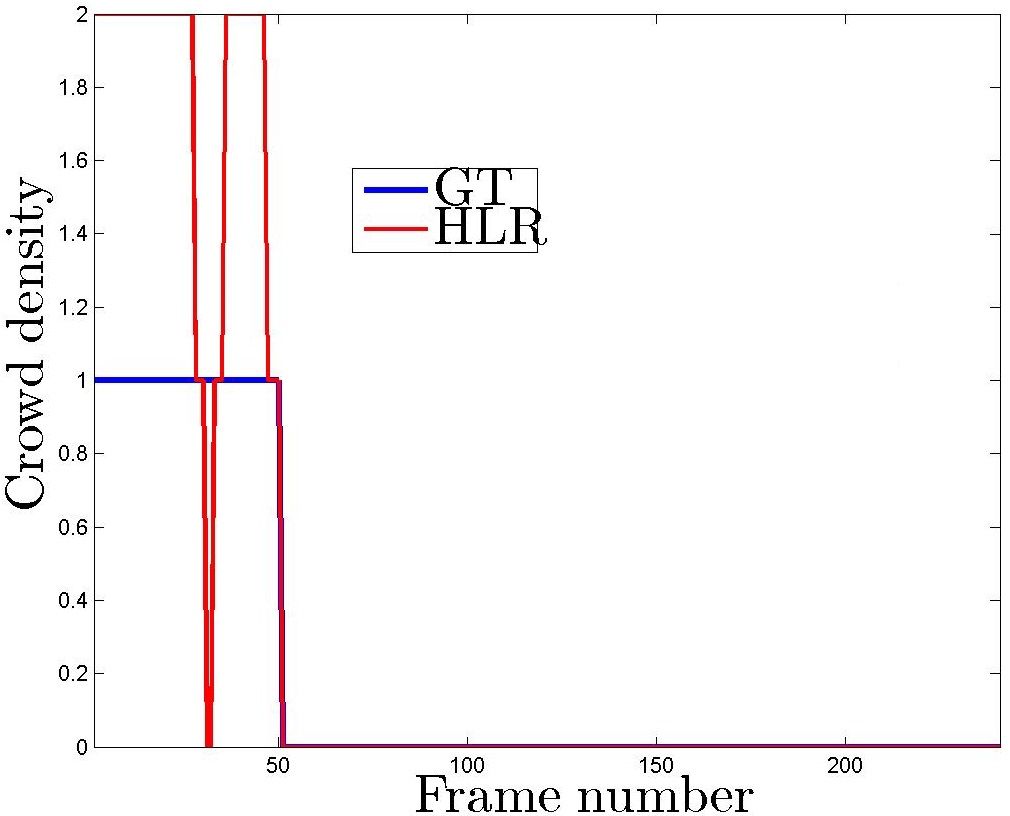}}%
	\subfigure[$13-59$ R1 total\label{g:s}]{\includegraphics[keepaspectratio, width=.2\textwidth]{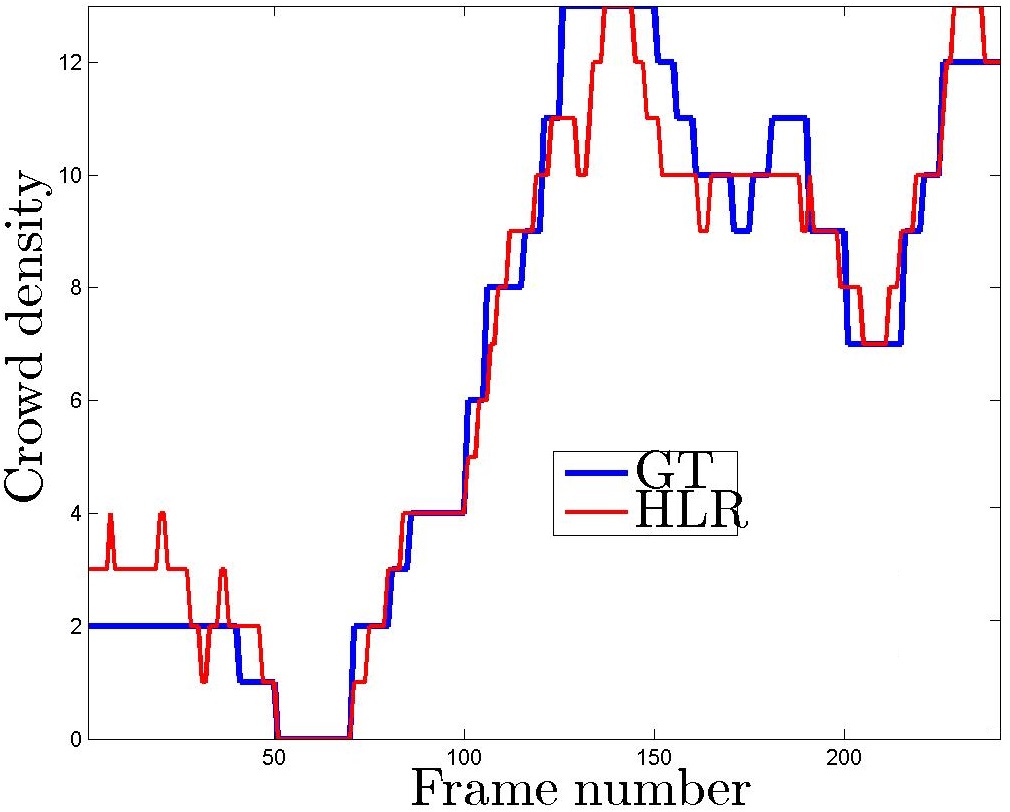}}%
	\caption{Qualitative results for HLR on crowd counting task. Ground truth crowd density (blue) is compared with HLR prediction (red). Graphs \ref{g:1} and \ref{g:3} refer to \emph{ UCSD} dataset, \ref{g:2} and \ref{g:4} to \emph{MALL}, all the others sequences are drawn from \emph{PETS 2009} dataset. Best viewed in color.}	
	\label{fig:qr1}
\end{figure*}

\emph{MALL} -- From a surveillance camera in a shopping centre, $2000$ RGB images were extracted (resolution $320 \times 240$). In each image, crowd density varies from $13$ to $53$. The main challenges are related to shadows and reflections. Following the literature \cite{Chen:BMVC12}, our system is trained with the first $800$ frames, and  the remaining ones are left for testing.

\emph{UCSD} -- A hand-held camera recorded a campus outdoor scene composed by $2000$ gray-scale $238$-by-$158$ frames. The density grows from $11$ to $46$. Environment changes are less severe, while geometric distorsion is sometimes a burden. As usually done \cite{Chan:CVPR2008}, we used the frames $601 \div 1400$ for training.

\emph{PETS 2009} -- Within the Eleventh Performance Evaluation of Tracking and Surveillance workshop, a new dataset has been recorded from a British campus. Crowd counting experiments are carried out on sequences $13$-$57$, $13$-$59$, $14$-$03$, $14$-$06$ from camera $1$ \cite{Chan:2009}, and three regions of interests have been introduced (R0, R1 and R2 in Fig. \ref{fig:data}). Crowd density ranges between $0$ and $42$, shadows and the appearance of both walking and running people are the main challenges. In Table \ref{tab:1}, the training/testing splits used \cite{Chan:2009}.

\begin{table}[t!]
	\centering
	\begin{tabular}{|r|c|c|c|c|c|}
		& \multicolumn{2}{c|}{Table \ref{tab:ryan}} & \multicolumn{2}{c|}{Table \ref{tab:semi} \& \ref{tab:multi}}  & Table \ref{tab:PETS} \\
		& \emph{UCSD}                & \emph{MALL}               & \emph{UCSD}                & \emph{MALL}  & \emph{PETS 2009}              \\ \hline\hline
		$T$                  & 0                   & 3                  & 4                   & 3     & 3             \\\hline
		$\lambda$            & $10^{-4}$           & $10^{-4}$          & $10^{-10}$          & $10^{-5}$    & $10^{-5}$      \\\hline
		$\gamma$             & $10^{-5}$           & $10^{-5}$          & $10^{-11}$          & $10^{-6}$     & $10^{-6}$     \\\hline
		$\Delta\xi$          & $0.10$              & $0.15$             & $0.05$              & $0.10$    & $0.10$ \\\hline       
	\end{tabular}
	\caption{Number of refinements $T$, regularizing parameter $\lambda$, $\gamma$ and rate $\Delta\xi$ used by HLR for crowd counting.}
	\label{tab:par}
\end{table}

In addition to replicating training/testing split, for a fair comparison, we employed publicly available\footnote{\url{http://personal.ie.cuhk.edu.hk/~ccloy/downloads_mall_dataset.html} for \emph{MALL}; \url{http://visal.cs.cityu.edu.hk/downloads/} for \emph{UCSD} and \emph{PETS 2009}.} ground truth annotations and pre-computed features: precisely, we employed \emph{size}, \emph{edges} and \emph{texture} features \cite{Ryan:2015}. Size descriptors refer to the magnitude of any interesting segments or patches from an image which are deemed to be relevant (e.g., the foreground pixels \cite{Davies:95}). 
Edges pertain to the relative changes in gray-level values and binary detectors (Canny algorithm \cite{Canny:86}) are used for extraction. 
The texture class includes several statistics, like energy or entropy, which are computed from gray-level co-occurrence matrix \cite{Haralick:73} or local binary pattern \cite{Gong}. Before extracting these descriptors, a region of interest is detected, perspective is normalized \cite{Ma:2004} and, sometimes, an intermediate motion segmentation phase is performed \cite{Chan:2009} (hence, crowd can be subdivided according to motion directions). 

In our framework, we set $m=3$ and each category of features is thus encoded with a separate (quadratic-polynomial or linear) kernel. We fix $c = [1/3,1/3,1/3]^\top$ and $M^\alpha$ is the sum of between-view operator from \cite{Minh:2013} and normalized graph Laplacian related to the $\alpha$-th view. The model parameters $T$, $\lambda$, $\gamma$ and $\Delta\xi$ are chosen via cross validation on the training set (see Table \ref{tab:par}).

\begin{figure*}[t!]	
	\centering	
	\subfigure[$13-59$ R2 left \label{s:a}]{\includegraphics[keepaspectratio, width=0.164\textwidth]{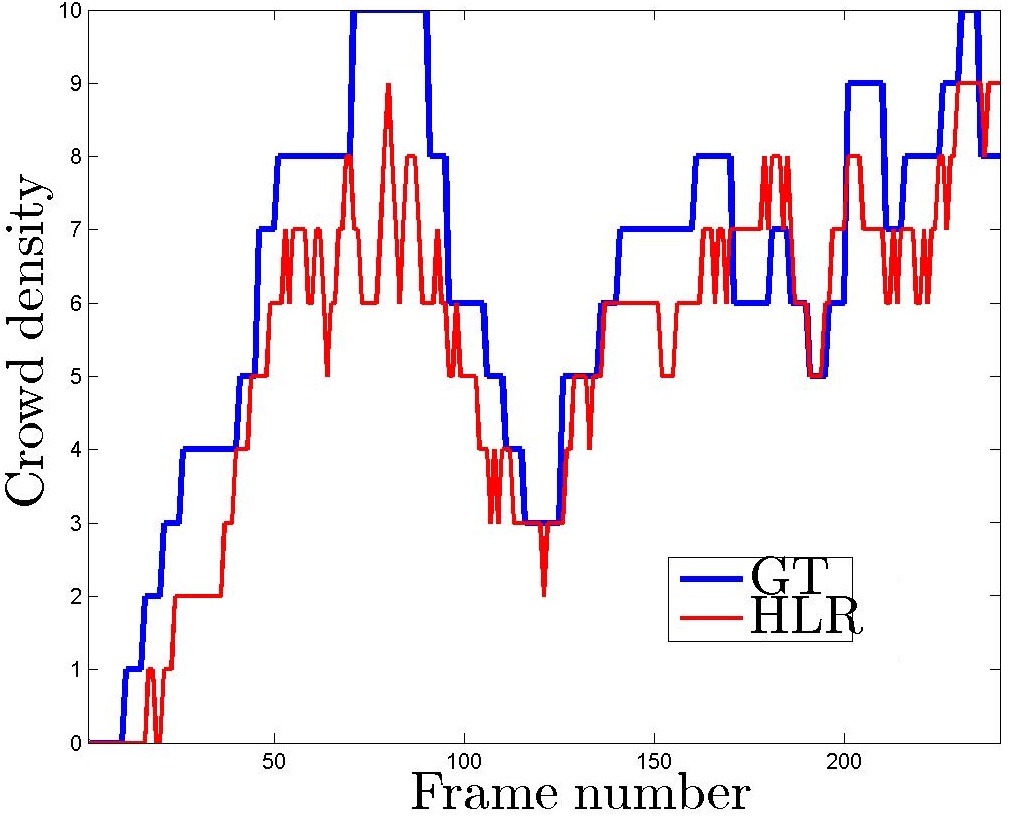}}%
	\subfigure[$13-59$ R2 right \label{s:b}]{\includegraphics[keepaspectratio, width=0.164\textwidth]{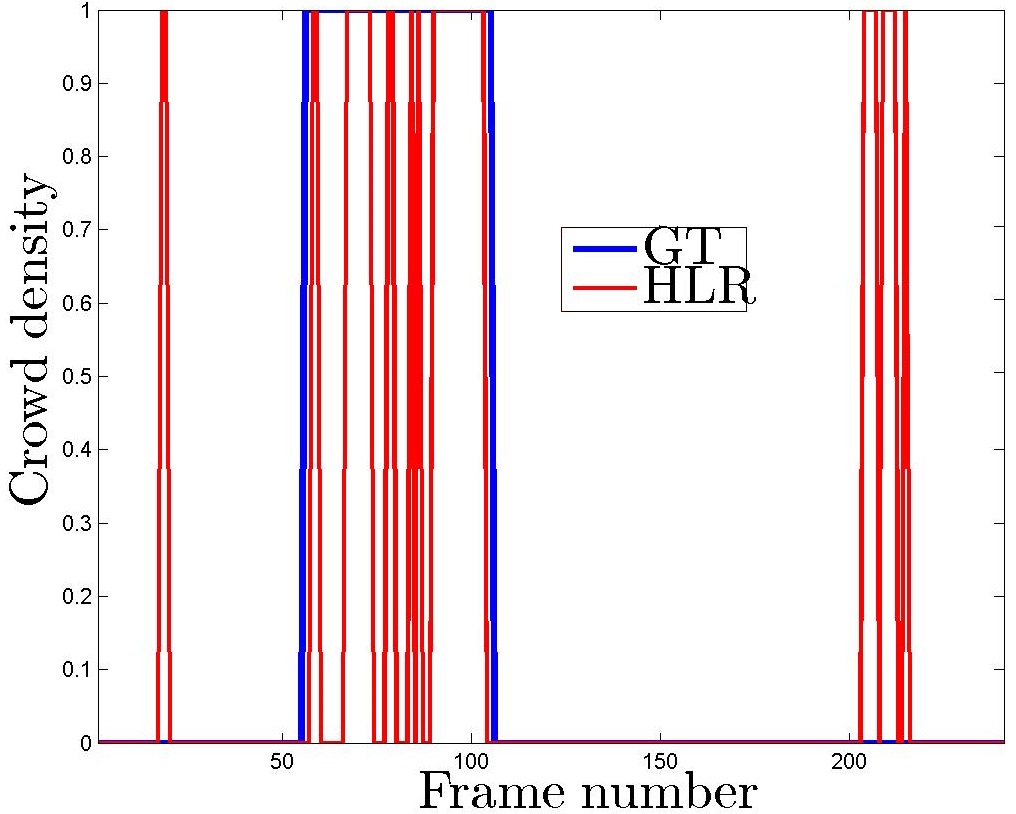}}%
	\subfigure[$13-59$ R2 total\label{s:c}]{\includegraphics[keepaspectratio, width=0.164\textwidth]{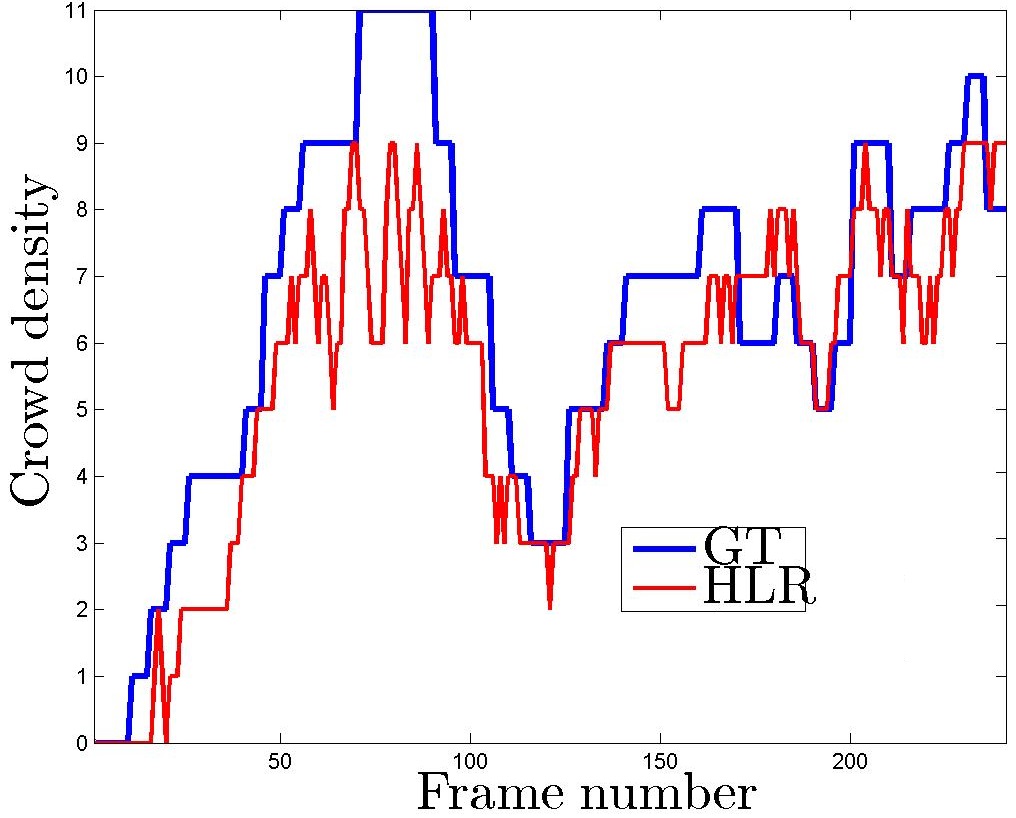}}%
	\subfigure[$14-06$ R1 left \label{s:d}]{\includegraphics[keepaspectratio, width=0.164\textwidth]{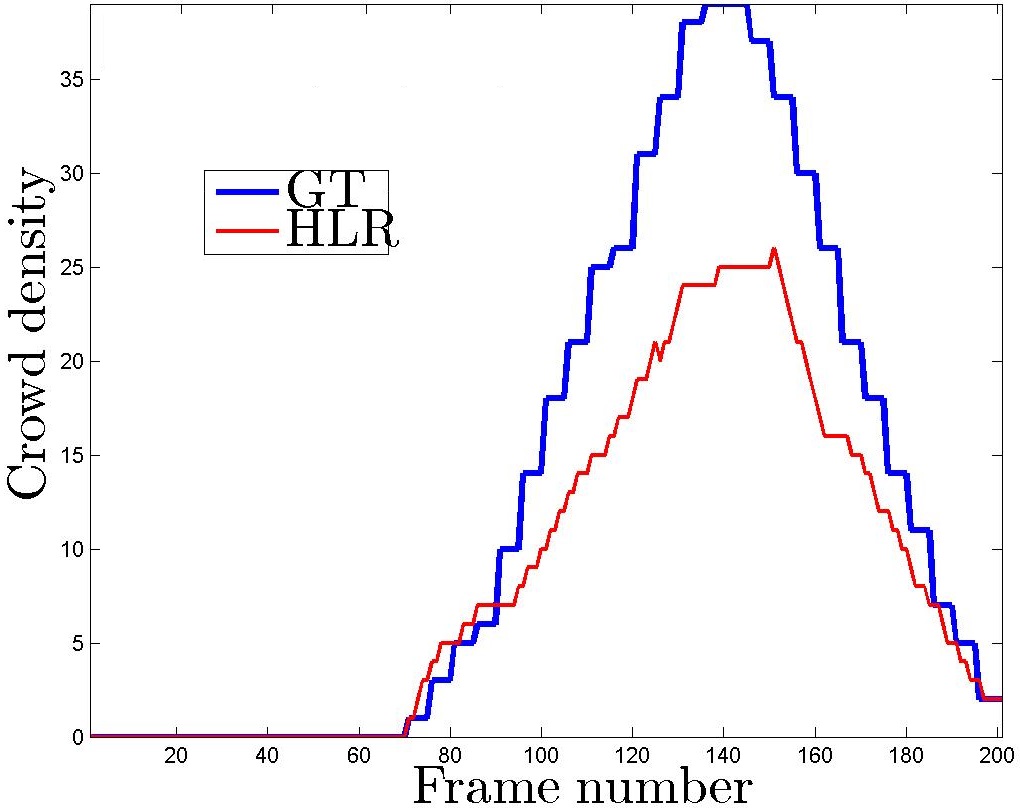}}%
	\subfigure[$14-06$ R1 right \label{s:e}]{\includegraphics[keepaspectratio, width=0.164\textwidth]{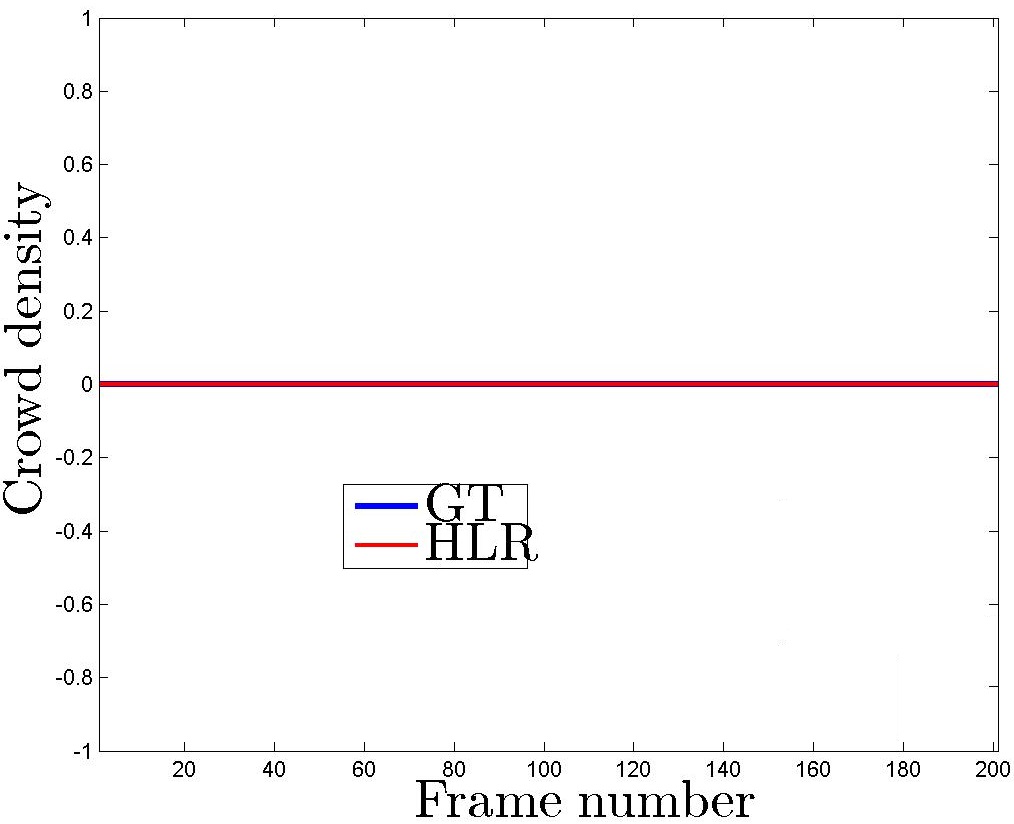}}%
	\subfigure[$14-06$ R1 total\label{s:f}]{\includegraphics[keepaspectratio, width=0.164\textwidth]{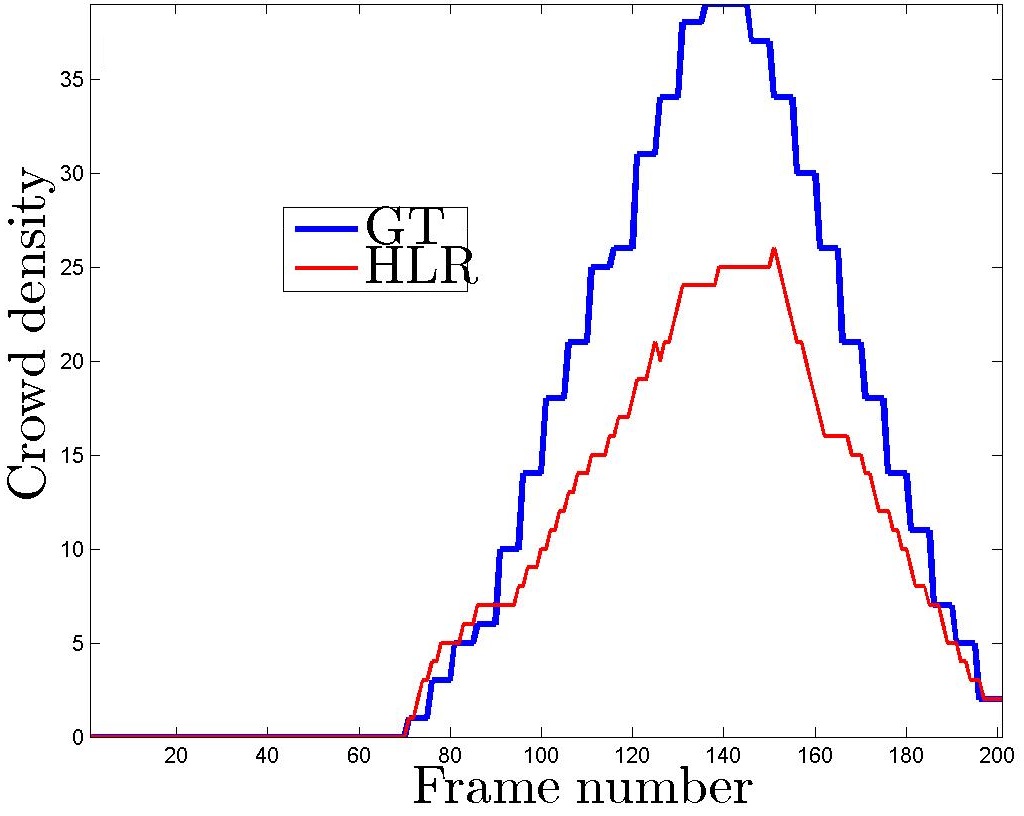}}\\
	\subfigure[$14-06$ R2 left \label{s:g}]{\includegraphics[keepaspectratio, width=0.164\textwidth]{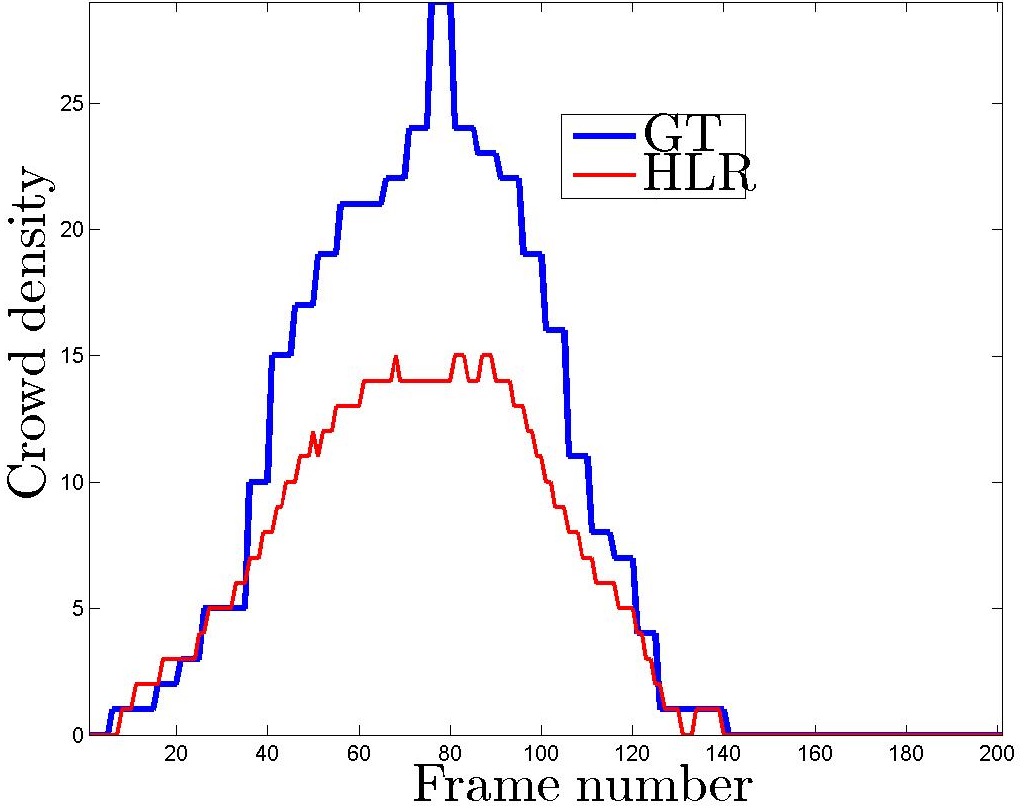}}%
	\subfigure[$14-06$ R2 right]{\includegraphics[keepaspectratio, width=0.164\textwidth]{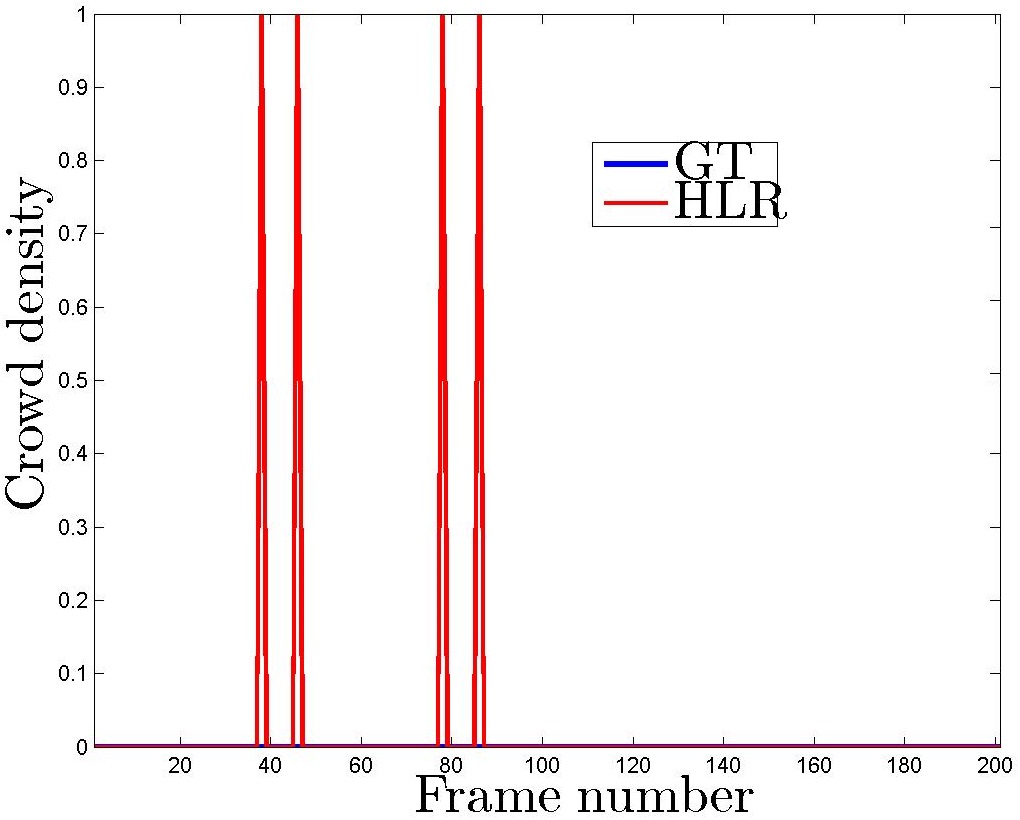}}%
	\subfigure[$14-06$ R2 total\label{s:i}]{\includegraphics[keepaspectratio, width=0.164\textwidth]{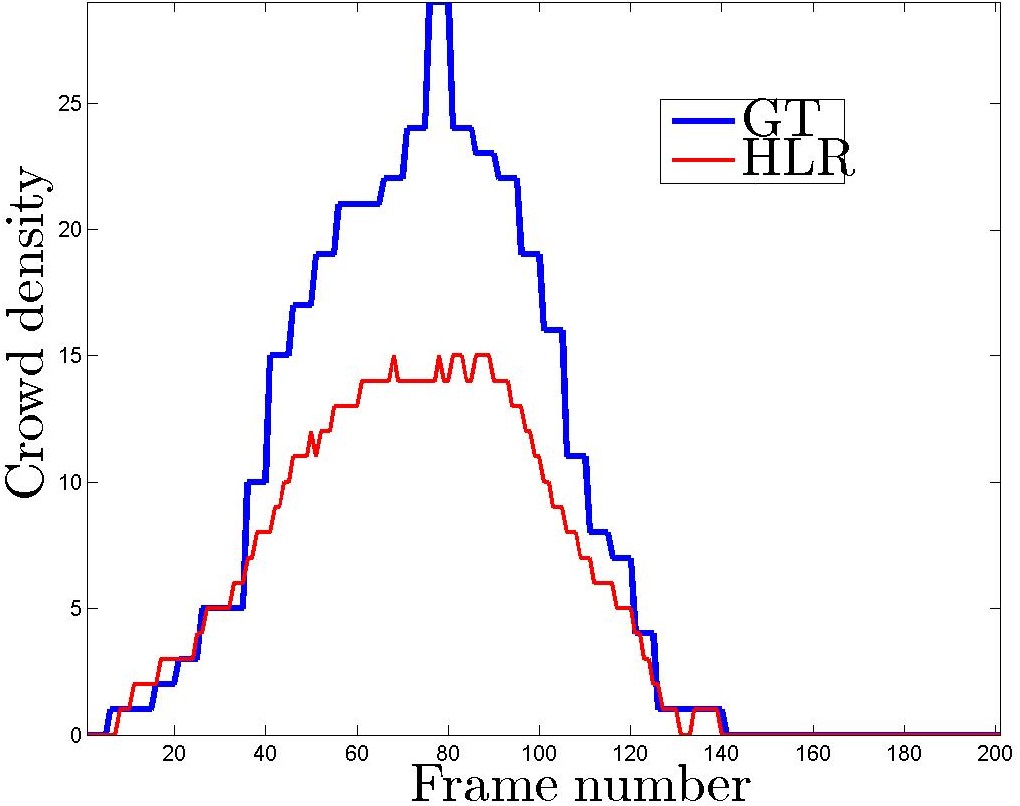}}%
	\subfigure[$14-17$ R1 left \label{s:j}]{\includegraphics[keepaspectratio, width=0.164\textwidth]{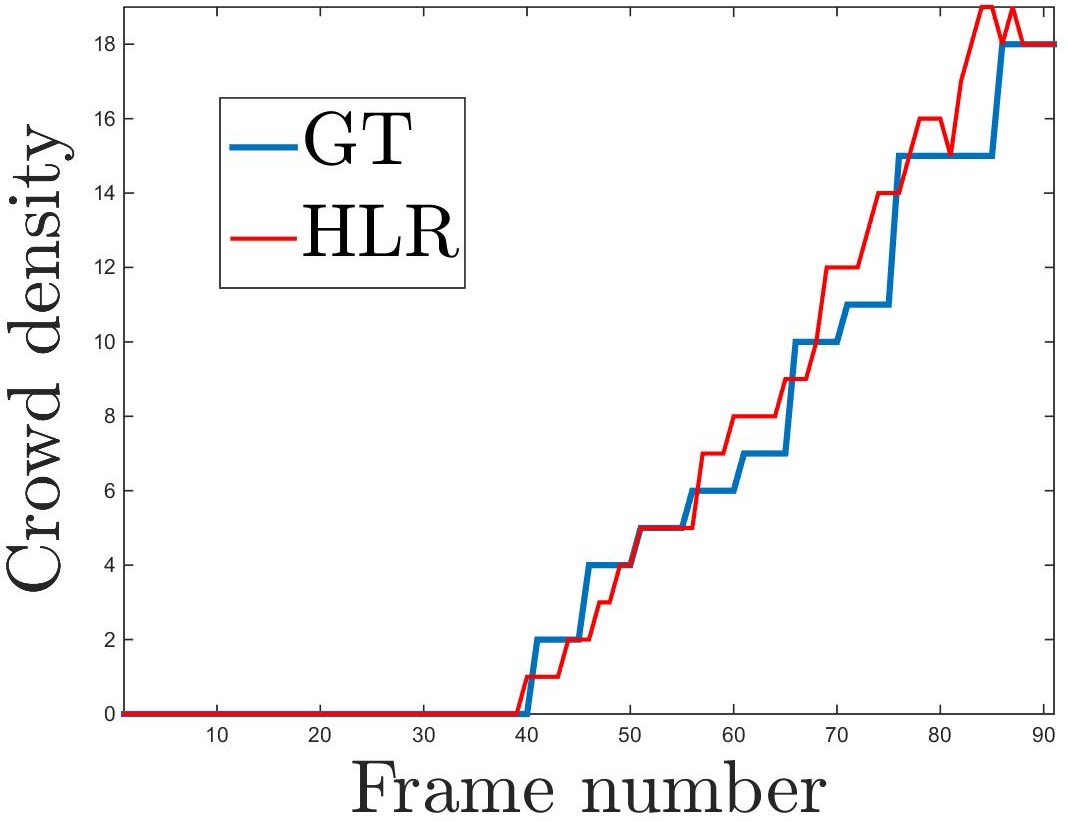}}%
	\subfigure[$14-17$ R1 right \label{s:k}]{\includegraphics[keepaspectratio, width=0.164\textwidth]{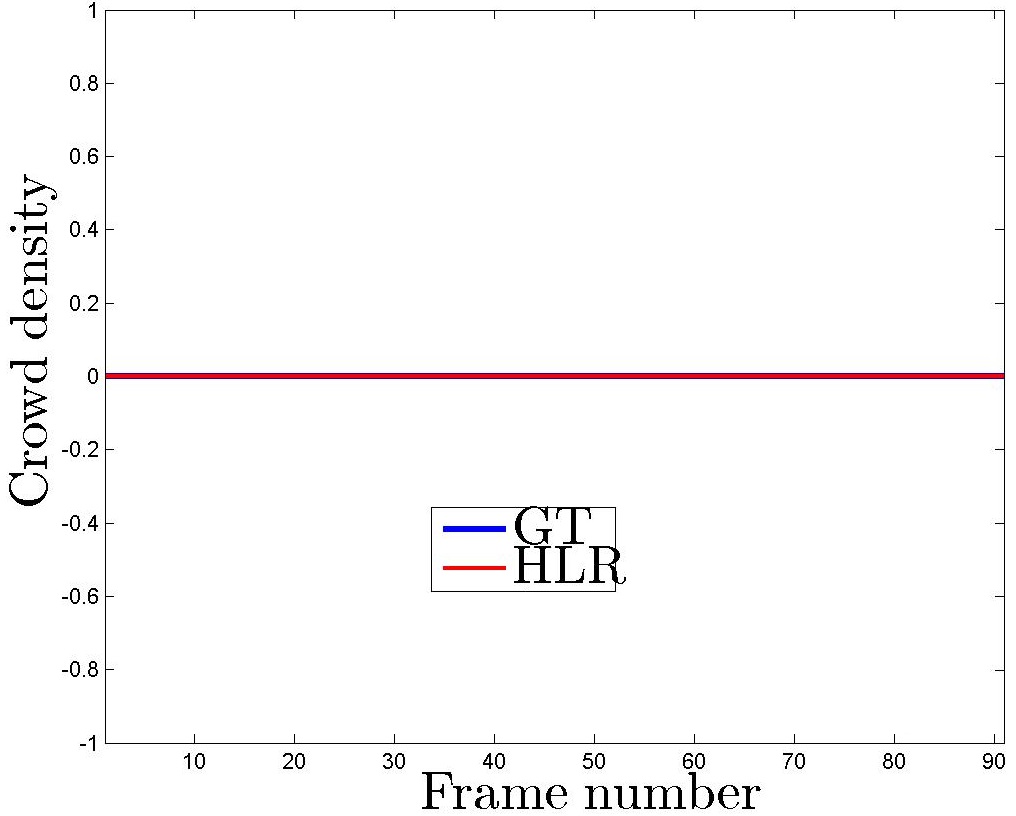}}%
	5\subfigure[$14-17$ R1 total \label{s:l}]{\includegraphics[keepaspectratio, width=0.164\textwidth]{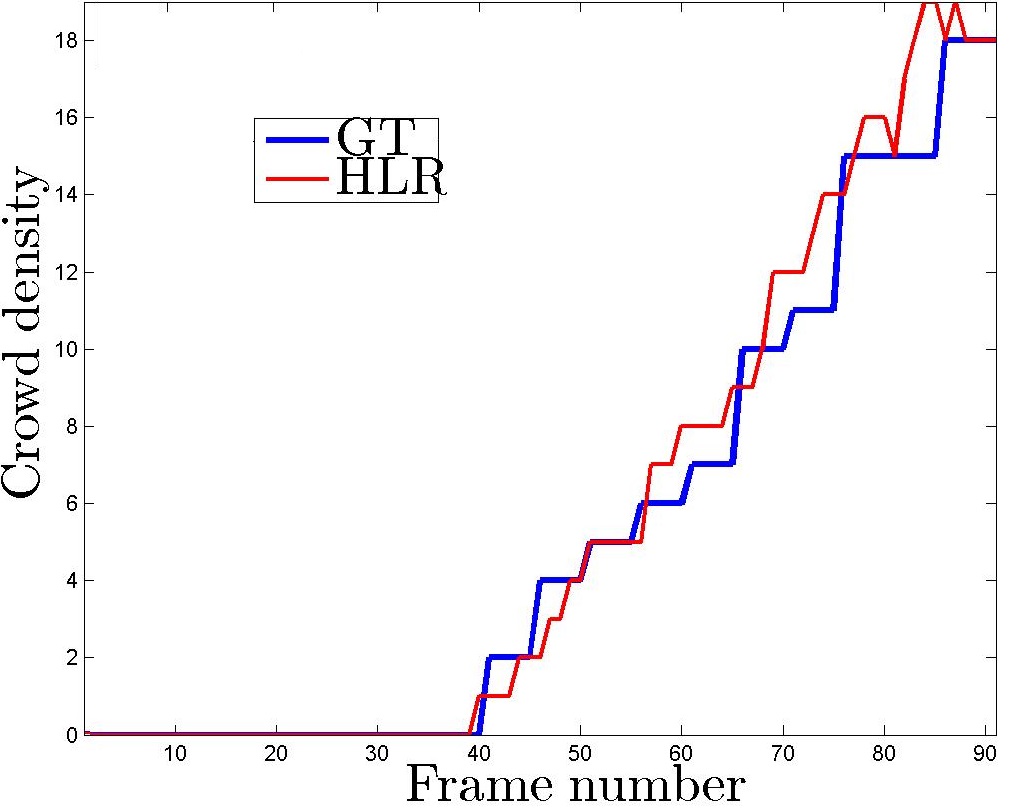}}%
	\caption{Qualitative results for HLR on \emph{PETS 2009} sequences. Blue: ground truth crowd density. Red: HLR prediction. Best viewed in color.}
	\label{fig:qr2}
\end{figure*}

Figures \ref{fig:qr1} and \ref{fig:qr2} show the qualitative results, overlapping the profile of the ground truth crowd density with our predicted evaluations for all the considered experiments: in all cases the ground truth crowd size is estimated in a satisfactory way. Quantitative performance evaluation uses $\MAE$, $\MSE$ and $\MRE$ metrics (Section \ref{sez:altri}).

Following the protocol of \cite{Ryan:2015}, in Table \ref{tab:ryan}, Huber Loss Regression (HLR) is compared with Gaussian Process Regression (GPR), regularized linear regression (Lin), $K$-nearest neighbours ($K$-nn) with $K=4$ and neural networks (NN) methods with a unique hidden layer composed by 8 artificial neurons. Using same training/testing splits as in \cite{Ryan:2015}, we scored top three error on \emph{MALL}, providing the lowest $\MAE$ and $\MRE$ on \emph{UCSD}.

\begin{table}[t!]
	\centering
	\begin{tabular}{|c|cc|cc|}
		Method & \multicolumn{2}{c|}{\emph{UCSD}} & \multicolumn{2}{c|}{\emph{MALL}} \\ 
		& $\MAE$ & $\MRE$ & $\MAE$ & $\MRE$ \\ \hline\hline
		GPR & {\bf 1.46}({\bf 2}) & {\bf 6.23}({\bf 2})  & {\bf 2.58}({\bf 1}) & {\bf 8.34}({\bf 1}) \\
		Lin & {\bf 1.56}({\bf 3}) & {\bf 6.48}({\bf 3}) & {\bf 2.58}({\bf 1}) & {\bf 8.52}({\bf 2}) \\
		$K$-nn & 2.72(4) & 9.63(4) & 2.89(4) & 9.28(4) \\
		NN & 9.15(5) & 43.08(5) & 13.02(5) & 43.40(5) \\\hline
		HLR & {\bf 1.23}({\bf 1}) & {\bf 5.43} ({\bf 1}) & {\bf 2.63}({\bf 3}) & {\bf 8.74}({\bf 3})\\\hline
	\end{tabular}
	\caption{Comparison using the protocol of \cite{Ryan:2015}. Top three lowest errors in bold, relative ranking in brackets.}\label{tab:ryan}
\end{table}


We also tested \emph{UCSD} and \emph{MALL} datasets in the same conditions as \cite{Chen:BMVC12} and \cite{Chen:CVPR2013} (see Table \ref{tab:multi}). Therein, the methods adopted in the comparison are least square support vector regression LSSVR \cite{Gestel:2001}, kernel ridge regression KRR \cite{An:2007}, random forest regression RFR \cite{Andy:2002}, Gaussian process regression GPR \cite{Chan:CVPR2008}, ridge regression RR \cite{Saunders:98} with its cumulative attribute CA-RR \cite{Chen:CVPR2013} variant. Additionally, we also compare with multiple localised regression MLR \cite{Wu:2006} and multiple output regression MORR \cite{Chen:BMVC12}, a class of local approaches for crowd counting which rely on a preliminary fine tessellation of the video frames. Also in this demanding comparison HLR is able to set the lowest and second lowest $\MAE,\MSE$ and $\MRE$ in all of the comparisons, respectively.

Additionally, we benchmarked HLR with other methods which leverage on semi-supervision: namely, the baseline one-viewed manifold regularization (MR) \cite{Belkin:2006}, semi-supervised-regression (SSR) \cite{Gong:ICCV13} and elastic net (EN) \cite{EN}. SSR optimizes a similar functional to \eqref{eq:probl}, where the quadratic loss is used in a multi-view setting ($m=2$) as to impose a spatial and temporal regularization within- and across-consecutive frames, respectively. EN \cite{EN} implements a sparsity principle while adopting a $L^1$-based semi-supervised variation of Lasso. In Table \ref{tab:semi} we report the $\MSE$ quantitative results, where HLR is able to outperform other semi-supervised methods.

\begin{table}[t!]
	\centering
	\begin{tabular}{|c|c|c|}
		Method & \emph{UCSD} & \emph{MALL} \\\hline\hline
		MR\cite{Belkin:2006} & 7.94(4) & 18.42(3) \\
		SSR\cite{Gong:ICCV13} & 7.06(3) & 17.85(2) \\
		EN\cite{EN} & {6.15}(2) & - \\\hline
		HLR & {\bf 6.00}({\bf 1}) & {\bf 16.42}({\bf 1}) \\\hline
	\end{tabular}
	\caption{Comparison with semi-supervised approaches. $\MSE$ error metric was used, relative ranking in brackets.}\label{tab:semi}
\end{table}

Moving to \emph{PETS 2009}, we mimed the protocol of \cite{Chan:2009}. Motion segmentation allows to divide the right-moving pedestrians from the others moving in the opposite direction. Total crowd density has been obtained summing the partial results. Table \ref{tab:PETS} shows a comparison of Huber loss vs. Gaussian Process Regression (GPR) in this setting reported in Table \ref{tab:1}. Performances are sometimes substantially improved, see sequence $13-57,$ regions R1 and R2. Again, HLR scored a sound performance, setting in 46 cases out of 54 the lowest $\MAE$ or $\MSE$ error metrics. 

{\bf Discussion.} $\bullet$ In comparison with the semi-supervised methods in Table \ref{tab:semi}, the considered multi-view and manifold regularized framework provides a better performance and Huber loss attests to be superior to both the quadratic (MR and SSR) and the $L^1$ losses (EN).

$\bullet$ The active-learning component is able to proficiently rule the amount of supervision. Indeed, on \emph{UCSD} only 1\% of the labels is not exploited by HLR: evidently, the preprocessing step perspective correction \cite{Chan:CVPR2008} is enough effective to make almost all the data exploitable in a supervised fashion. Differently, on \emph{MALL}, about 11\% of labelled instances are discarded: this happens when some pedestrians are partially occluded by some static elements of the scene and, sometimes, there are some sitting people whose appearance greatly differs from the walking ones. Finally, on \emph{PETS 2009}, HLR outperforms GPR even if using, on average, more than 100 annotations less. Despite using less labelled data than competitors, HLR scores a superior performance on \emph{UCSD}, \emph{MALL} and \emph{PETS 2009} datasets.

$\bullet$ Similarly to Section \ref{sez:altri}, HLR performance does not requires any burdensome parameter tuning: a good rule-of-the-thumb is $T=3,$ $\lambda = 10^{-4}$, $\gamma = 10^{-5}$, $\Delta \xi = 0.01$ and, in general, only minor corrections are required (Table \ref{tab:par}).

$\bullet$ In terms of running time, the HLR is a fast method: indeed, in the setup of Table \ref{tab:multi}, training and testing on \emph{MALL} last 6.5 and 0.4 seconds respectively. Similarly, on \emph{UCSD}, training requires 5.6 and testing 0.5 seconds.

$\bullet$ In synthesis, the crowd counting application showed that HLR is able to fully take advantage of the most effective techniques in semi-supervision and to improve state-of-the-art methods, while being robust to noisy annotations, ensuring a fast computation and skipping annoying parameter-tuning processes.

	\begin{table*}[t!]
		\centering
		\begin{tabular}{|c|ccc|ccc|}
			Method & & \emph{UCSD} & & & \emph{MALL} & \\ 
			& $\MAE$ & $\MSE$ & $\MRE$ & $\MAE$ & $\MSE$ & $\MRE$ \\ \hline\hline
			LSSVR\cite{Gestel:2001} & 2.20(4) & 7.69(4) &  {\bf 0.107(3)}  & 3.51(4) &  18.20(5) & 0.108(4) \\
			KRR\cite{An:2007} & {\bf 2.16}({\bf 3}) & {\bf 7.45}({\bf 3}) & {\bf 0.107(3)}  &  3.51(4) &  18.18(4) &  0.108(4) \\
			RFR\cite{Andy:2002} & 2.42(8) &  8.47(8) &  0.116(8)  & 3.91(9) & 21.50(8) & 0.121(9) \\
			GPR\cite{Chan:CVPR2008} &  2.24(5) & 7.97(6) &  0.112(7)  & 3.72(7) & 20.10(7) & 0.115(7) \\
			RR\cite{Saunders:98} & 2.25(6) & 7.82(5) & 0.110(6) & 3.59(6) & 19.00(6) & 0.110(6) \\
			CA-RR\cite{Chen:CVPR2013} & {\bf 2.07}({\bf 2}) & {\bf 6.86}({\bf 2}) & {\bf 0.102}({\bf 2}) & {\bf 3.43}({\bf 3}) & {\bf 17.70}({\bf 3}) & {\bf 0.105}({\bf 3}) \\
			MLR\cite{Wu:2006} &  2.60(9) & 10.10(9) & 0.125(9)  & 3.90(8) & 23.90(9) & 0.120(8) \\
			MORR\cite{Chen:BMVC12} & 2.29 (7) & 8.08(7) & 0.109(5)  & {\bf 3.15}({\bf 1}) &  {\bf 15.70}({\bf 1}) & {\bf 0.099}({\bf 1}) \\\hline
			HLR & {\bf 1.99} ({\bf 1}) & {\bf 6.00}({\bf 1}) & {\bf  0.093}({\bf 1}) & {\bf 3.36}({\bf 2}) & {\bf 16.42}({\bf 2}) &  {\bf 0.104}({\bf 2})\\\hline
		\end{tabular}
		\caption{Comparison of HLR using the protocol of \cite{Chen:BMVC12} and \cite{Chen:CVPR2013}. Top three performance in bold, relative ranking in brackets.}\label{tab:multi}
	\end{table*}

	\begin{table*}[t!]
		\centering
		\begin{tabular}{|ccc|cc|cc|cc|}
			Sequence & Region & Method &		\multicolumn{2}{c|}{total} & \multicolumn{2}{c|}{right-moving} & \multicolumn{2}{c|}{left-moving} \\ 
			& & & $\MAE$ & $\MSE$ & $\MAE$ & $\MSE$ & $\MAE$ & $\MSE$ \\ \hline\hline
			$13$-$57$ & R0 & GPR & 2.308 & 8.362 & 0.249 & 0.339 & 2.475 & 8.955 \\
			& & HLR & {\bf 2.290} & {\bf 8.118} & {\bf 0.204} & {\bf 0.204} & {\bf 2.385} & {\bf 8.719} \\ \hline
			$13$-$57$ & R1 & GPR & 1.697 & 5.000 & 0.100 & 0.100 & 1.643 & 4.720 \\
			& & HLR & {\bf 1.330} & {\bf 3.005} & {\bf 0.059} & {\bf 0.059} & {\bf 1.290} & {\bf 2.919} \\ \hline
			$13$-$57$ & R2 & GPR & 1.072 & 1.796 & 0.235 & 0.317 & 0.842 & 1.484 \\
			& & HLR & {\bf 0.819} & {\bf 1.253} &  {\bf 0.081} &  {\bf 0.081} & {\bf 0.756} & {\bf 1.190} \\ \hline
			$13$-$59$ & R0 & GPR & 1.647 & 4.087 & 1.668 & 4.158 & 0.154 & 0.154 \\
			& & HLR & {\bf 1.560} & {\bf 3.320} & {\bf 1.639} & {\bf 3.589} & {\bf 0.137} & {\bf 0.137} \\ \hline
			$13$-$59$ & R1 & GPR & 0.685 & 1.116 & 0.589 & 0.871 & {\bf 0.095} & {\bf 0.095} \\
			& & HLR & {\bf 0.622} & {\bf 0.855} & {\bf 0.481} & {\bf 0.689} &  0.166 & 0.166 \\ \hline
			$13$-$59$ & R2 & GPR & 1.282 & {\bf 2.577} & 1.291 & 2.436 & {\bf 0.066} & {\bf 0.066} \\		
			& & HLR & {\bf 1.253} & 2.747 & {\bf 1.195} & {\bf 2.274} &  0.141 & 0.141 \\ \hline
			$14$-$06$ & R1 & GPR & 4.328 & 44.159 & 4.338 & 44.159 & 0.005 & 0.005 \\
			& & HLR & {\bf 4.299} & {\bf 43.383} & {\bf 4.299} & {\bf 43.383} & {\bf 0.000} & {\bf 0.000} \\ \hline
			$14$-$06$ & R2 & GPR & 3.139 & 26.035 & 3.144 & 26.129 & 0.020 & 0.020 \\
			& & HLR & {\bf 2.995} & {\bf 23.970} & {\bf 3.015} &  {\bf 24.289} & {\bf 0.020} & {\bf 0.020} \\ \hline
			$14$-$17$ & R1 & GPR & 0.604 & 1.220 & 0.604 & {\bf 1.198} & {\bf 0.000} & {\bf 0.000} \\
			& & HLR & {\bf 0.593} & {\bf 1.209} & {\bf 0.593} &  1.209 &  {\bf 0.000} & {\bf 0.000} \\\hline
		\end{tabular}
		\caption{Comparison of HLR with the protocol of \cite{Chan:2009} on \emph{PETS 2009}. The lowest error is in bold. Sometimes a 0.000 error value is registered: it correspond to the absence of people moving in the specified direction in the given sequence.}\label{tab:PETS}
	\end{table*}

\section{Conclusion, Limitations \& Future Work}\label{sez:end}

In this paper we provide a novel adaptive solution to exactly minimize the Huber loss in a general optimization framework which unifies the most effective approaches in semi-supervision (multi-view learning and manifold regularization). Differently from previous approaches \cite{Mangasarian:00, Ando:05, LL:11,  Khan:13}, the proposed HLR algorithm $1)$ avoids burdensome iterative solving while, at the same time, $2)$ automatically learns the threshold $\xi$ and $3)$ actively selects the most beneficial annotations for the learning phase. 

Such unique aspects resulted in a remarkable performance on different tasks where HLR scored always on par and often superior to several state-of-the-art algorithms for learning with noisy annotations, classical regression problems and crowd counting application. Moreover, low errors were registered by HLR at low time complexity, guaranteeing a quick computation without requiring fine parameter-tuning.

Future works will essentially focus on adapting the framework for vector-valued regression. Also, to face the main drawback of the method, which consists in the not-learnt weights $c$, the connections with M-estimators theory, Multiple Kernel Learning and Radial Basis Function Networks could be investigated.


%

%
%

\ifCLASSOPTIONcaptionsoff
  \newpage
\fi



\bibliographystyle{IEEEtran}
\bibliography{fonti}

\appendix

\section{Proof of the main theorem}\label{sez:app}

In this Section we report the proof of Theorem \ref{th:th} while applying the Representer Theorem \cite{Minh:2011} to cast the optimization problem into a minimizing on the coefficients $\mathbf{w}$ defining $f^\star$ in \eqref{eq:exp}. Precisely, implementing it in \eqref{eq:probl} yields 
\begin{align}
\mathcal{J}_{\lambda,\gamma}(\mathbf{w}) &= \dfrac{1}{\ell} \sum_{i=1}^{\ell} H_\xi\left( y_i - \sum_{j=1}^{u+\ell} c^\top K(\mathbf{x}_i,\mathbf{x}_j)w_j\right) + \nonumber\\ &+\lambda \sum_{j,k=1}^{u+\ell} w_j^\top K(\mathbf{x}_j,\mathbf{x}_k) w_k + \label{eq:zero} \\ &+\gamma \sum_{i,j=1}^{u+\ell}\sum_{h,k=1}^{u+\ell} w_h^\top K(\mathbf{x}_h,\mathbf{x}_i)\textsc{M}_{ij}K(\mathbf{x}_j,\mathbf{x}_k)w_j. \nonumber
\end{align}   
Theoretically, minimizing $J_{\lambda,\gamma}$ over the RKHS $\mathcal{S}_K$ is fully equivalent to minimizing $\mathcal{J}_{\lambda,\gamma}$ with respect to $\mathbf{w},$ being the latter approach computationally convenient because, for this purpose, the optimization domain $\mathbb{R}^{m(u+\ell)}$ is preferable to an infinite-dimensional functional space. Notice that each addend of $\mathcal{J}_{\lambda,\gamma}$ is differentiable with respect to $\mathbf{w}$. Indeed, one has
\begin{equation}\label{eq:H'}
H'_\xi(y) = \begin{cases} - \xi & \mbox{if} \; y \leq -\xi \\ y & \mbox{if} \; |y| \leq \xi \\ +\xi & \mbox{if} \; y \geq \xi,\end{cases}
\end{equation}
and $\|f\|_K^2$ and $\|f\|^2_M$ in \eqref{eq:zero} are also differentiable since polynomials in  $w_1,\dots,w_{u+\ell}.$ Thus, for any $p=1,\dots,u+\ell$ and $\eta = 1,\dots,m,$ we compute $\dfrac{\partial \mathcal{J}_{\lambda,\gamma}}{\partial w_p^\eta},$ differentiating with respect to $w_p^\eta,$ the $\eta$-th view  of $w_p.$ Then,
\begin{align}
\dfrac{\partial \mathcal{J}_{\lambda,\gamma}}{\partial w_p^\eta} = -\dfrac{1}{\ell} \sum_{i=1}^\ell H'_\xi\left(y_i - \sum_{j=1}^{u+\ell} c^\top K(\mathbf{x}_i,\mathbf{x}_j) w_j\right) \cdot \nonumber\\ \cdot \sum_{h=1}^{u+\ell}\sum_{\alpha=1}^m c^\alpha \kappa^\alpha(x^\alpha_i,x^\alpha_h)\dfrac{\partial w_h^\alpha}{\partial w_p^\eta} + \nonumber\\ \lambda \sum_{j,k=1}^{u+\ell} \sum_{\alpha=1}^m\dfrac{\partial w_j^\alpha}{\partial w_p^\eta}\kappa^\alpha(x^\alpha_j,x^\alpha_k)w^\alpha_k + \nonumber \\ \lambda \sum_{j,k=1}^{u+\ell} \sum_{\alpha=1}^m w^\alpha_j\kappa^\alpha(x^\alpha_j,x^\alpha_k)\dfrac{\partial w_k^\alpha}{\partial w_p^\eta} +  \nonumber \\ \gamma \sum_{i,j=1}^{u+\ell}\sum_{h,k=1}^{u+\ell} \sum_{\alpha=1}^m \dfrac{\partial w_h^\alpha}{\partial w_p^\eta} \kappa^\alpha(x^\alpha_i,x^\alpha_h)M_{ij}^{\alpha}\kappa^\alpha(x^\alpha_j,x^\alpha_k)w_k^\alpha  + \nonumber \\  \gamma \sum_{i,j=1}^{u+\ell}\sum_{h,k=1}^{u+\ell} \sum_{\alpha=1}^m  \kappa^\alpha(x^\alpha_i,x^\alpha_h)w^\alpha_hM_{ij}^{\alpha}\kappa^\alpha(x^\alpha_j,x^\alpha_k)\dfrac{\partial w_k^\alpha}{\partial w_p^\eta}.\label{eq:prima}
\end{align}

In order to simplify \eqref{eq:prima}, we can apply the relationship
$\dfrac{\partial w^\alpha_i}{\partial w^\eta_j} = \delta_{\alpha\eta}\delta_{ij},$ valid for any $\alpha,\beta = 1,\dots,m$ and $i,j=1,\dots,u+\ell,$ where, for any integers $m,n,$ $\delta_{mn}$ is the Kronecker delta and $\delta_{mn} = 1$ if $m=n,$ while, otherwise, $\delta_{mn} = 0$ if $m \neq n.$ Thus,

\begin{align}
\dfrac{\partial \mathcal{J}_{\lambda,\gamma}}{\partial w_p^\eta} = -\dfrac{1}{\ell} \sum_{i=1}^\ell H'_\xi\left(y_i - \sum_{j=1}^{u+\ell} c^\top K(\mathbf{x}_i,\mathbf{x}_j) w_j\right) \cdot \nonumber\\ \cdot \sum_{h=1}^{u+\ell}\sum_{\alpha=1}^m c^\alpha \kappa^\alpha(x^\alpha_i,x^\alpha_h)\delta_{\alpha\eta}\delta_{hp} + \nonumber\\ \lambda \sum_{j,k=1}^{u+\ell} \sum_{\alpha=1}^m\delta_{\alpha\eta}\delta_{jp}\kappa^\alpha(x^\alpha_j,x^\alpha_k)w^\alpha_k + \nonumber \\ \lambda \sum_{j,k=1}^{u+\ell} \sum_{\alpha=1}^m w^\alpha_j\kappa^\alpha(x^\alpha_j,x^\alpha_k)\delta_{\alpha\eta}\delta_{kp} +  \nonumber \\ \gamma \sum_{i,j=1}^{u+\ell}\sum_{h,k=1}^{u+\ell} \sum_{\alpha=1}^m \delta_{\alpha\eta}\delta_{hp} \kappa^\alpha(x^\alpha_i,x^\alpha_h)M_{ij}^{\alpha}\kappa^\alpha(x^\alpha_j,x^\alpha_k)w_k^\alpha  + \nonumber \\ \gamma \sum_{i,j=1}^{u+\ell}\sum_{h,k=1}^{u+\ell} \sum_{\alpha=1}^m  \kappa^\alpha(x^\alpha_i,x^\alpha_h)w^\alpha_hM_{ij}^{\alpha}\kappa^\alpha(x^\alpha_j,x^\alpha_k)\delta_{\alpha\eta}\delta_{kp}.\label{eq:taleggio}
\end{align}
By exploiting the properties of Kronecker delta, inside a summation over the index $i$, $\delta_{ij}$ discards all the addends except to $j.$ Then, we can rewrite equation  \eqref{eq:taleggio} obtaining
\begin{align}
\hspace{-0.6cm}\dfrac{\partial \mathcal{J}_{\lambda,\gamma}}{\partial w_p^\eta} = -\dfrac{1}{\ell} \sum_{i=1}^\ell H'_\xi\left(y_i - \sum_{j=1}^{u+\ell} c^\top K(\mathbf{x}_i,\mathbf{x}_j) w_j\right)  \cdot \nonumber \\ \cdot c^\eta \kappa^\eta(x^\eta_i,x^\eta_p) + \nonumber \\ \lambda \sum_{k=1}^{u+\ell} \kappa^\eta(x^\eta_p,x^\eta_k)w^\eta_k + \lambda \sum_{j=1}^{u+\ell} w^\eta_j\kappa^\eta(x^\eta_j,x^\eta_p) +  \nonumber \\ \gamma \sum_{i,j=1}^{u+\ell}\sum_{k=1}^{u+\ell} \kappa^\eta(x^\eta_i,x^\eta_p)M_{ij}^{\eta}\kappa^\eta(x^\eta_j,x^\eta_k)w_k^\eta  + \nonumber \\ \gamma \sum_{i,j=1}^{u+\ell}\sum_{h=1}^{u+\ell}   \kappa^\eta(x^\eta_i,x^\eta_h)w^\eta_hM_{ij}^{\eta}\kappa^\eta(x^\eta_j,x^\eta_p).\label{eq:seconda}
\end{align}

To rearrange equation \eqref{eq:seconda}, we can exploit the functional symmetry of both Mercer kernels $\kappa^1,\dots,\kappa^m$ and linear operators $M^1,\dots,M^m$. Then, one sees 
\begin{align}
\hspace{-0.6cm}\dfrac{\partial \mathcal{J}_{\lambda,\gamma}}{\partial w_p^\eta} = -\dfrac{1}{\ell} \sum_{i=1}^\ell H_\xi'\left(y_i - \sum_{j=1}^{u+\ell} c^\top K(\mathbf{x}_i,\mathbf{x}_j) w_j\right) \cdot \nonumber \\ \cdot  c^\eta \kappa^\eta(x^\alpha_i,x^\alpha_p) + \nonumber\\ 2\lambda \sum_{k=1}^{u+\ell} w^\eta_k\kappa^\eta(x^\alpha_p,x^\alpha_k) + \nonumber \\ 2\gamma \sum_{i,j=1}^{u+\ell}\sum_{k=1}^{u+\ell} \kappa^\eta(x^\alpha_i,x^\alpha_p)M_{ij}^{\eta}\kappa^\eta(x^\alpha_j,x^\alpha_k)w_k^\eta. \label{eq:terza}
\end{align}

After vectorizing with respect to $\eta = 1,\dots,m,$ the derivative $\dfrac{\partial \mathcal{J}_{\lambda,\gamma}}{\partial w_p}$ equals to
\begin{align}
-\dfrac{1}{\ell} \sum_{i=1}^\ell H_\xi'\left(y_i - \sum_{j=1}^{u+\ell} c^\top K(\mathbf{x}_i,\mathbf{x}_j) w_j\right)  K(\mathbf{x}_p,\mathbf{x}_i) c  + \nonumber \\ 2\lambda \sum_{k=1}^{u+\ell} K(\mathbf{x}_p,\mathbf{x}_k)\omega_k + \nonumber \\ 2\gamma \sum_{i,j=1}^{u+\ell}\sum_{k=1}^{u+\ell} K(\mathbf{x}_p,\mathbf{x}_i)\textsc{M}_{ij}K(\mathbf{x}_j,\mathbf{x}_k)w_k \label{eq:quarta}. 
\end{align}
Expression \eqref{eq:quarta} rewrites $\sum_{i=1}^{u+\ell} K(\mathbf{x}_p,\mathbf{x}_i)\psi_i,$ once defined, for any $i=1,\dots,u+\ell,$
\begin{align}
\hspace{-.4cm}\psi_i = -\mathds{1}( i\leq \ell) \dfrac{1}{\ell} H_\xi'\left(y_i - \sum_{j=1}^{u+\ell} c^\top K(\mathbf{x}_i,\mathbf{x}_j) w_j\right) c + \nonumber \\ 2\lambda\omega_i + 2\gamma \sum_{j,h=1}^{u+\ell}\textsc{M}_{ij}K(\mathbf{x}_j,\mathbf{x}_h)w_h, \label{eq:sesta}
\end{align}
where the indicator function $\mathds{1}$ is conditionally defined to be $\mathds{1}( i\leq \ell) = 1$ if $i \leq \ell$ and  $\mathds{1}( i\leq \ell) = 0$ if $i > \ell.$ 

If we set $\psi_1=\dots=\psi_{u+\ell}=0,$ then, from equation \eqref{eq:terza}, $\dfrac{\partial \mathcal{J}_{\lambda,\gamma}}{\partial w_1}=\dots=\dfrac{\partial \mathcal{J}_{\lambda,\gamma}}{\partial w_{u+\ell}}=0$ and this leads to a solution of \eqref{eq:probl}. But, this is the only solution we have since, as motivated in the paper, the optimization problem \eqref{eq:probl} has unique solution thanks to Representer Theorem \cite{Minh:2011}. Then, the previous discussion ensures that, globally, the two systems of equations are totally equivalent since, for every $i = 1,\dots,u+\ell,$
\begin{equation}
\psi_i = 0 \quad \mbox{if and only if} \quad \dfrac{\partial \mathcal{J}_{\lambda,\gamma}}{\partial w_i} = 0.
\end{equation}
Hence, the optimization of \eqref{eq:probl} can be done by solving
\begin{align}
2\ell \lambda w_i + 2 \ell \gamma \sum_{j,h=1}^{u+\ell} \textsc{M}_{ij} K(\mathbf{x}_j,\mathbf{x}_h)w_h = \nonumber \\ H_\xi'\left(y_i - \sum_{j=1}^{u+\ell}c^\top K(\mathbf{x}_i,\mathbf{x}_j)w_j\right)c \label{eq:Lu}
\end{align} for $i=1,\dots,\ell;$ and, when $i=\ell+1,\dots,u+\ell,$ 
\begin{equation}\label{eq:Uu}
2\lambda w_i + 2 \gamma \sum_{j,h=1}^{u+\ell} \textsc{M}_{ij}K(\mathbf{x}_j,\mathbf{x}_h)w_h =0.
\end{equation}
Substitute equation \eqref{eq:H'} into \eqref{eq:Lu}. Then, for $i = 1,\dots,\ell,$ 
\begin{align}
2\ell \lambda w_i + 2 \ell \gamma \sum_{j,h=1}^{u+\ell} \textsc{M}_{ij} K(\mathbf{x}_j,\mathbf{x}_h)w_h = \nonumber \\
\begin{cases} - \xi c \\ & \hspace{-3.5 cm} \mbox{if} \; y_i - \sum_{j=1}^{u+\ell}c^\top K(\mathbf{x}_i,\mathbf{x}_j)w_j \leq -\xi \\ \left(y_i - \displaystyle \sum_{j=1}^{u+\ell}c^\top K(\mathbf{x}_i,\mathbf{x}_j)w_j\right)c \\ & \hspace{-3.5 cm} \mbox{if} \; \left|y_i - \sum_{j=1}^{u+\ell}c^\top K(\mathbf{x}_i,\mathbf{x}_j)w_j\right| \leq \xi \\ +\xi c \\ & \hspace{-3.5 cm} \mbox{if} \; y_i - \sum_{j=1}^{u+\ell}c^\top K(\mathbf{x}_i,\mathbf{x}_j)w_j \geq \xi.\end{cases}\label{eq:Lu'}
\end{align}
If one defines the following set of indices
\begin{align*}  
L_+[\mathbf{D},\mathbf{w},\xi] &= \left\{ i \leq \ell \colon \sum_{j=1}^{u+\ell} c^\top K(\mathbf{x}_i,\mathbf{x}_j)w_j \geq y_i + \xi \right\}\hspace{-.1cm},\\ 
L_0[\mathbf{D},\mathbf{w},\xi] &= \left\{ i \leq \ell \colon \left|\sum_{j=1}^{u+\ell} c^\top K(\mathbf{x}_i,\mathbf{x}_j)w_j - y_i \right| < \xi \right\}\hspace{-.1cm},\\		 
L_-[\mathbf{D},\mathbf{w},\xi] &= \left\{ i \leq \ell \colon \sum_{j=1}^{u+\ell} c^\top K(\mathbf{x}_i,\mathbf{x}_j)w_j \leq y_i - \xi \right\}\hspace{-.1cm},\\ 
\end{align*} 
equation \eqref{eq:Lu'} therefore becomes
\begin{align}
& 2\ell \lambda w_i + 2 \ell \gamma \sum_{j,h=1}^{u+\ell} \textsc{M}_{ij} K(\mathbf{x}_i,\mathbf{x}_h)w_h = \label{eq:Lu''} \nonumber \\ &\begin{cases}
-\xi c & \mbox{if} \; i \in L_+[\mathbf{D},\mathbf{w},\xi] \\
\left(y_i - \displaystyle \sum_{j=1}^{u+\ell} c^\top K(\mathbf{x}_i,\mathbf{x}_j) w_j \right) c & \mbox{if} \; i \in L_0[\mathbf{D},\mathbf{w},\xi] \\
+\xi c & \mbox{if} \; i \in L_-[\mathbf{D},\mathbf{w},\xi].
\end{cases}
\end{align}
The thesis follows as the straightforward combination of equations \eqref{eq:Lu''} and \eqref{eq:Uu}.

\end{document}